\documentclass[journal]{IEEEtran}

\usepackage{times}
\usepackage{epsfig}
\usepackage{graphicx}
\usepackage{amsmath}
\usepackage{amssymb}
\usepackage{lineno}
\usepackage{multirow}
\usepackage{caption}
\usepackage{epstopdf}
\usepackage{subcaption}
\usepackage{hyperref}
\usepackage[table,xcdraw]{xcolor}
\DeclareMathOperator*{\argmin}{argmin}
\ifCLASSINFOpdf
\else
\fi
\begin{document}
\title{Visual Saliency Detection Based on Multiscale Deep CNN Features}

\author{Guanbin Li and Yizhou Yu
% <-this % stops a space
\thanks{G. Li is with Sun Yat-sen University, Guangzhou 510006, China, and also with the Department of Computer Science, the University of Hong Kong (e-mail: ligb86@gmail.com).}
\thanks{
 Y. Yu is with the Department of Computer Science, the University of Hong Kong (e-mail: yizhouy@acm.org).}
% <-this % stops a space
\thanks{A preliminary version of this paper appeared in CVPR 2015~\cite{li2015visual}.}% <-this % stops a space
\thanks{Project website \color{blue}\href{url}{https://sites.google.com/site/ligb86/mdfsaliency/}}}


\markboth{}%
{}



\maketitle

\begin{abstract}
Visual saliency is a fundamental problem in both cognitive and computational sciences, including computer vision. In this paper, we discover that a high-quality visual saliency model can be learned from multiscale features extracted using deep convolutional neural networks (CNNs), which have had many successes in visual recognition tasks. For learning such saliency models, we introduce a neural network architecture, which has fully connected layers on top of CNNs responsible for feature extraction at three different scales.
The penultimate layer of our neural network has been confirmed to be a discriminative high-level feature vector for saliency detection, which we call deep contrast feature. To generate a more robust feature, we integrate handcrafted low-level features with our deep contrast feature.
To promote further research and evaluation of visual saliency models, we also construct a new large database of 4447 challenging images and their pixelwise saliency annotations. Experimental results demonstrate that our proposed method is capable of achieving state-of-the-art performance on all public benchmarks, improving the F-measure by 6.12\% and 10.0\% respectively on the DUT-OMRON dataset and our new dataset (HKU-IS), and lowering the mean absolute error by 9\% and 35.3\% respectively on these two datasets.
\end{abstract}


% \fi
\IEEEpeerreviewmaketitle

\section{Introduction}
\IEEEPARstart{V}isual saliency attempts to determine the amount of attention steered towards various regions in an image by the human visual and cognitive systems~\cite{BI13}. It is thus a fundamental problem in psychology, neural science, and computer vision. Computer vision researchers focus on developing computational models for either simulating the human visual attention process or identifying visually salient regions. It is originally defined as a task of predicting eye-fixations to investigate the mechanism of human visual system~\cite{itti1998model}. Recently it has been extended to locating regions of interest, known as salient object detection\cite{liu2011learning,achanta2009frequency}. Since visual saliency results set relative importance on the visual contents in an image, they are conducive to narrowing the scope of visual processing and saving computing resources. As a result,
Visual saliency has been incorporated in a variety of computer vision and image processing tasks to improve their performance. Such tasks include image cropping~\cite{Autocollage}, retargeting~\cite{SeamCarving}, summarization~\cite{SCSI08} and thumbnail generation~\cite{MCC09}. Recently, visual saliency has also been increasingly used by visual recognition tasks%~\cite{RWKP04}
, such as object tracking~\cite{wu2014weighted}, image classification~\cite{WYW13} and person re-identification~\cite{bi2014person}.

\begin{figure}[ht]
\begin{center}
   \includegraphics[width=0.45\textwidth]{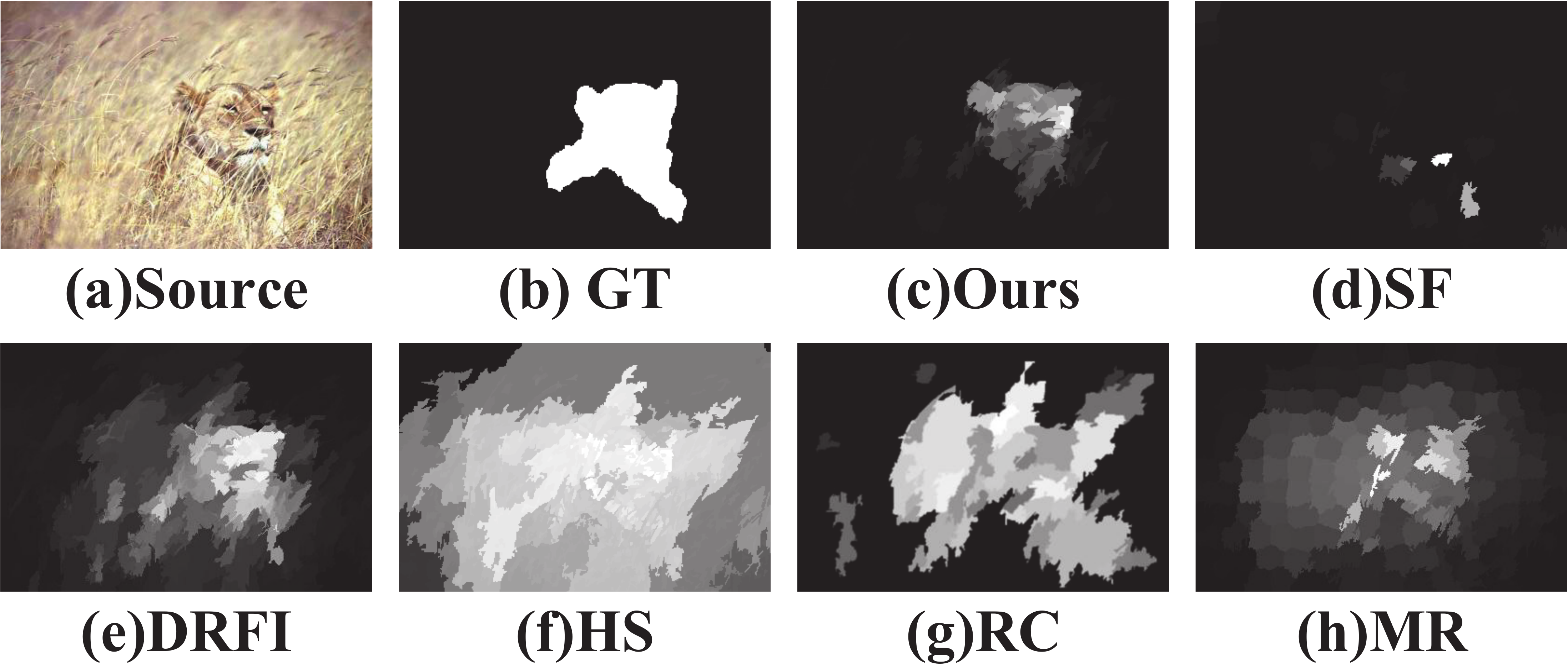}
\end{center}
   \caption{An example illustrating that saliency models based on handcrafted low-level features are fragile. From top left to bottom right: source image, ground truth, our saliency map, and saliency maps of other five latest methods, including SF\cite{perazzi2012saliency}, DRFI\cite{jiang2013salient}, HS\cite{yan2013hierarchical}, RC\cite{cheng2011global}, and MR\cite{yang2013saliency}.}
\label{fig:low-contrast}
\end{figure}

Results from perceptual research~\cite{itti2001computational,reinagel1999natural} show that contrast is the most influential factor to visual attention in the human vision system. Local and global contrast has been successfully adopted to derive saliency maps in various saliency detection methods, where the definition of contrast is based on various types of handcrafted image features (e.g., color, intensity and histogram) at the pixel or superpixel level~\cite{cheng2011global,liu2011learning,yang2013saliency}. Though these methods perform well on simple benchmarks, they may fail when the background becomes complex since handcrafted low-level features are not able to effectively capture semantic contexts hidden in an image, and very often the contrast between these low-level features is not strong enough to make salient objects stand out from the background. For example, in Figure \ref{fig:low-contrast}, a lion is hidden in the bushes and it could not be detected as a salient object using low-level saliency cues alone. However, humans can easily recognize the lion and check it out carefully since it is semantically salient in high-level cognition. Because of this, in our work, we leverage the advantages of high-level semantically meaningful features from deep learning as well as low-level features when inferring saliency maps.

Human visual and cognitive systems involved in the visual attention process are composed of layers of interconnected neurons. For example, the human visual system has layers of simple and complex cells whose activations are determined by the magnitude of input signals falling into their receptive fields. Since deep artificial neural networks were originally inspired by biological neural networks, it is a natural choice to build a computational model of visual saliency using deep artificial neural networks. Specifically, recently popular convolutional neural networks (CNN) are particularly well suited for this task because convolutional layers in a CNN resemble simple and complex cells in the human visual system~\cite{fukushima1980neocognitron} while fully connected layers in a CNN act like higher-level inference and decision making. % of our human brain which perceives an overall consideration of all observed things.
%things from global information which resemble higher-level inference and decision making in the human cognitive system.

In this paper, we develop a new computational model for visual saliency using multiscale deep features computed by convolutional neural networks. Deep neural networks, such as CNNs, have recently achieved many successes in visual recognition tasks~\cite{krizhevsky2012imagenet,FCNL13,girshick2014rich}.
%Given a large set of training images, a deep CNN with a huge number of parameters can be trained to accurately grasp the essential concepts and knowledge from the dataset. For example, there are 60 million parameters and 650,000 neurons in a CNN trained on the ImageNet image database~\cite{deng2009imagenet}.
Such deep networks are capable of extracting feature hierarchies from raw pixels automatically. Further, features extracted using such networks are highly versatile and often more effective than traditional handcrafted features. Inspired by this, we perform feature extraction using a CNN originally trained over the ImageNet dataset~\cite{deng2009imagenet}. Since ImageNet contains images of a large number of object categories, our features contain rich semantic information, which is useful for visual saliency because humans pay varying degrees of attention to objects from different semantic categories.
%These features have turned out to be very effective for visual saliency computation in our experiments. This is because these pre-trained networks are capable of extracting not only low-level image features related to color and contrast but also discriminative semantic features as ImageNet contains images of a large number of object categories. Such semantic features are useful for visual saliency because humans pay varying degrees of attention to objects from different semantic categories.
For example, viewers of an image likely pay more attention to objects like cars than the sky or grass. In the rest of this paper, we call such features {\em CNN features}.

By definition, saliency is resulted from visual contrast as it intuitively characterizes certain parts of an image that appear to stand out relative to their neighboring regions or the rest of the image.
%Thus, to generate a saliency map, we run our saliency model repeatedly over a set of image regions.
Thus, to compute the saliency of an image region, our model should be able to evaluate the contrast between the considered region and its surrounding area as well as the rest of the image. Therefore, we extract multiscale CNN features for every image region from three nested and increasingly larger rectangular windows, which respectively encloses the considered region, its immediate neighboring regions, and the entire image.

On top of the multiscale CNN features, our method further trains fully connected neural network layers. Concatenated multiscale CNN features are fed into these layers trained using a collection of labeled saliency maps. Thus, these fully connected layers play the role of a regressor that is capable of inferring the saliency score of every image region from the multiscale CNN features extracted from nested windows surrounding the image region. %It is well known that deep neural networks with at least one fully connected layers can be trained to achieve a very high level of regression accuracy.
The penultimate fully connected layer of our neural network is thus becoming a very discriminative high-level feature vector for saliency detection, and we can generate significantly more accurate saliency maps than those from existing saliency models based on low-level features by simply performing logistic regression. We further find out that this high-level discriminative feature vector is complementary to handcrafted low-level features, and train a random forest regressor on concatenated high-level and low-level features. %We concatenate it with several low-level features extracted as proposed in \cite{jiang2013salient}, which can explicitly reflect the contrast and spatial information around the saliency region. The concatenated feature vector is further fed into a random forest regressor which maps the feature vector of each region to a saliency score, and
Experimental results show that such hybrid features can further boost the performance of saliency detection.

We have extensively evaluated our CNN-based visual saliency model over existing datasets, and meanwhile noticed a lack of large and challenging datasets for training and testing saliency models. At present, MSRA-B~\cite{liu2011learning} is the most frequently used dataset. %the only large dataset that can be used for training a deep neural network based model was derived from the MSRA-B dataset~\cite{liu2011learning}.
However, this dataset has become less challenging over the years because images there typically include a single salient object located away from the image boundary. DUT-OMRON~\cite{yang2013saliency} is currently the most challenging dataset with nature images for the research of both salient object detection and eye fixation prediction. %Though DUT-OMRON~\cite{yang2013saliency} is another large dataset, many saliency annotations in this dataset may be controversial among different human observers. 
To facilitate research and evaluation of advanced saliency models, we have created another large dataset where an image likely contains multiple salient objects, which have a more general spatial distribution in the image. Furthermore, our dataset only includes images that receive consistent saliency annotations from multiple users.
%Since a saliency model runs over image regions instead of an entire image, all image regions discovered from a labeled image can serve as training samples. Thus, the size of our region-based training set is about two orders of magnitude larger than the collection of labeled images.
Our proposed saliency model has significantly outperformed all existing saliency models over this new dataset as well as all existing datasets.

In summary, this paper has the following contributions:
\begin{itemize}
\item A new visual saliency model is proposed to incorporate multiscale CNN features extracted from nested windows with a deep neural network with multiple fully connected layers.
    %The CNN features are extracted using a CNN originally trained over the ImageNet dataset.
    The deep neural network for saliency estimation is trained using regions from a set of labeled saliency maps. The penultimate layer of the proposed neural network can be viewed as a discriminative high-level feature vector for saliency detection, and can further boost saliency performance when concatenated with handcrafted low-level features.
\item A complete saliency framework is developed by further integrating an aggregated saliency map over multi-level image segmentations with a spatial coherence model based on a fully connected CRF.

    %Hierarchical image segmentation generates multiscale image regions, the initial saliency of which is estimated using our CNN-based model.
    %In our framework, saliency scores from our CNN-based model are first refined using the spatial coherence model, and further fused together across multiple levels of segmentation.
%\item A new challenging dataset, HKU-IS, is created for saliency model research and evaluation.
   %In this dataset, there are typically multiple salient objects in each image and these salient objects have a more general spatial distribution.
%This dataset is publicly available.
%Our proposed saliency model has been successfully validated on this new dataset as well as on all existing datasets.
\end{itemize}

The remainder of the paper is organized as follows. Section~\ref{sec:relatedwork} reviews related work and differentiates our method from such work. Section~\ref{sec:deep} introduces our proposed multiscale deep features. The complete algorithm is presented in Section~\ref{sec:algo}. A new dataset was introduced in the preliminary version of this paper~\cite{li2015visual}, we present it here again in Section~\ref{sec:data} for the completeness of this paper. Extensive experimental results and comparisons
are presented in Section~\ref{sec:experiment}. And Section~\ref{sec:conclusion} concludes this paper.

\section{Related Work}\label{sec:relatedwork}
\subsection{Salient Object Detection}
Visual saliency algorithms can be categorized into three groups: bottom-up, top-down, and hybrid algorithms of the previous two.

Bottom-up models are primarily based on the center-surround scheme, computing a master saliency map using a linear or non-linear combination of low-level visual attributes such as color, intensity, texture and orientation~\cite{itti1998model,hou2007saliency,achanta2009frequency,ChengPAMI,liu2011learning}. According to the spatial scope of saliency computation, these methods can be further divided into local methods and global methods. Local methods measure saliency by considering the contrast between each pixel or image region and a small neighborhood. One example of this category is the work by Itti~{\em et al.}~\cite{itti1998model}, where color and orientation contrasts across multiple scales are computed to measure local conspicuity. While it is able to identify salient pixels, as pointed out by Cheng~{\em et al.}~\cite{cheng2011global}, the results are generally blurry and contain a significant amount of false detection. Ma and Zhang~\cite{ma2003contrast} proposed a fuzzy growing process to simulate the process of human perception using local contrast as a measure of saliency.
Harel~{\em et al.}~\cite{harel2006graph} created feature maps using the method from \cite{itti1998model} but perform normalization using graph-based random walk. As these methods only consider local contrast, they tend to detect high-frequency features, such as edges or noise, only and suppress homogeneous regions at the interior of salient objects.

Global bottom-up methods estimate saliency by considering contrast over the entire image. Achanta~\cite{achanta2009frequency} proposed a frequency-tuned method that directly estimates pixel saliency by computing color differences from the average image color. Cheng~{\em et al.}~\cite{cheng2011global,ChengPAMI} took color histograms as regional features and computed saliency on the basis of histogram dissimilarity. %To account for spatial relationships inside an image, Perazzi~{\em et al.}~\cite{perazzi2012saliency} derived a saliency measure from the uniqueness and spatial distribution of certain compact and perceptually homogeneous elements.
In \cite{yan2013hierarchical}, Yan~{\em et al.} proposed a hierarchical framework to address small-scale high-contrast patterns. Recently, much effort has been made towards designing discriminative features and saliency priors. Most algorithms essentially follow the region contrast framework, aiming to discover features that better characterize the distinctiveness of an image region with respect to its surrounding area. In \cite{liu2011learning}, three novel features are integrated with a conditional random field. A model based on low-rank matrix recovery is presented in \cite{shen2012unified} to integrate low-level visual features with higher-level priors. Chen~{\em et al.}~\cite{chen2015structure} designs a structure-aware descriptor based on the intrinsic biharmonic distance metric which is able to simultaneously integrate local and global structure information. %The wavelet transform is employed in \cite{imamoglu2013saliency} to create multiscale features that modulate local contrast with global saliency. 
Though significant improvements have been made, these global features are still weak in capturing image semantic information.

Top-down methods in general require the incorporation of high-level knowledge, such as objectness and object detectors in the computational process~\cite{jia2013category,chang2011fusing,shen2012unified}. In \cite{judd2009learning}, Judd trained a top-down saliency model using high-level image features including those based on face detection and person detection results. Borji~{\em et al.}~\cite{borji2012boosting} integrated bottom-up and top-down features when learning their saliency model, considering person and car detectors as high-level priors. %Yang {\em et al.}~\cite{yang2012top} proposed a top-down saliency model that jointly learns a conditional random field and a discriminative dictionary. 
In \cite{jia2013category}, Jia~{\em et al.} computed a high-level saliency prior using objectness without category information, and applied a Gaussian MRF to enforce the consistency among salient regions. Chang~{\em et al.}~\cite{chang2011fusing} proposed a framework which conceptually integrates objectness and saliency via a graphical model accounting for their relationship. Our deep feature extracted from Krizhevsky's CNN~\cite{krizhevsky2012imagenet} implicitly encodes the semantic information of 1.2 million images and has much stronger generalization capability than those based on a relatively small number of object detectors (e.g. face, human and car) or approximate objectness.

Saliency priors, such as the center prior~\cite{liu2011learning,judd2009learning} and the boundary prior~\cite{jiang2013salient,zhu2014saliency}, are widely used to heuristically improve saliency estimation. The center prior is normally formulated as a Gaussian fall-off map assigning higher saliency to the central region of an image while the boundary prior takes a complementary perspective and assigns image boundary regions lower saliency.
These saliency priors are either directly integrated with other saliency cues as weights~\cite{ChengPAMI,cheng2013efficient,jia2013category} or used as features in learning based algorithms~\cite{jiang2013salient,judd2009learning}. While these empirical priors can improve saliency results for many images, they can fail when a salient object is off-center or significantly overlaps with the image boundary. Note that object location cues and boundary-based background modeling are not neglected in our framework, but have been implicitly incorporated through multiscale CNN feature extraction and neural network training.

\subsection{Deep Convolutional Neural Networks}
Convolutional neural networks have recently achieved many successes in visual recognition tasks, including image classification~\cite{krizhevsky2012imagenet}, object detection~\cite{girshick2014rich}, and scene parsing~\cite{FCNL13}. Donahue~{\em et al.}~\cite{donahue2014decaf} pointed out that features extracted from Krizhevsky's CNN trained on the ImageNet dataset~\cite{deng2009imagenet} can be repurposed to generic tasks. Razavian~{\em et al.}~\cite{sharif2014cnn} extended their results and concluded that CNN-based deep learning can be a strong candidate for any visual recognition tasks. Nevertheless, saliency detection is generally defined as a low-level computer vision problem and acts quite different from conventional object detection. %It   cannot be directly solved using the framework from \cite{donahue2013decaf,razavian2014cnn}. 
It is the contrast against the surrounding area rather than the content inside an image region that should be learned for saliency prediction. This paper proposes a simple but very effective neural network architecture for digging out contrast information hidden in multi-scale deep CNN features and inferring the saliency score for each region. Note that in~\cite{FCNL13}, a multiscale convolutional network was trained to extract hierarchical feature vectors well suited for scene labeling. The raw input image was transformed through a Laplacian pyramid into three scales before being fed to a 3-stage convolutional network, and the pixelwise features are similar to hypercolumn features~\cite{hariharan2015hypercolumns}, formed by stacking responses corresponding to the same pixel from all convolutional layers of the CNN. Different from region-oriented features used in our method, their pixel-oriented features are not focused on region contrast which is crucial in saliency detection.

There exist other convolutional neural network based saliency detection methods since the publication of our earlier work~\cite{li2015visual}. Wang~{\em et al.}~\cite{wang2015deep} applied a deep neural network (DNN-L) to learn local patch features for determining the saliency score of the center pixel. Since only local patches were considered, the quality of the generated saliency map may be sensitive to high-frequency background noise, and homogeneous regions inside salient objects may be misclassified. Therefore, a global search stage was added to exploit the complex relationships among global saliency cues which are represented using handcrafted features. Li~{\em et al.}~\cite{LiYu16} proposes an end-to-end deep contrast network which considers both pixel-level and segment-wise saliency inference. In~\cite{zhao2015saliency}, both global and local contexts were utilized and integrated into a unified deep learning framework for saliency detection. Their model calculates a saliency score for every superpixel. The global context of a superpixel contains the whole image with the superpixel located at the center of the context, while the local context has a fixed size equal to one third of the global context. While our proposed method also extracts CNN-based context features, it is different from \cite{zhao2015saliency} in three aspects and is also more robust. First, the size of our local context is spatially varying, relying on the actual size of the surrounding regions. Our local context can better estimate the contrast between each region and the background.
%Considering the case when multiple similar salient objects scatter near each other, context scope with the same size may include two or more scatter saliency objects, thus may bring in some ambiguity and decay the model.
Second, Instead of direct regression, we propose a neural network architecture to mine the contrast information hidden inside the concatenated multiscale deep features. %rather than applying direct regression.
Third, we apply multi-level segmentation and pixel-level CRF-based refinement to compensate the inaccuracy caused by superpixels. Experimental results demonstrate that our proposed method outperforms all existing CNN based saliency models.

This paper provides a more complete understanding of multiscale deep features first presented in the conference version~\cite{li2015visual}, providing additional insights, analysis, and evaluation. Furthermore, we improve the original framework in two aspects. First, we propose the concept of deep contrast features, and analyze their strengths and weaknesses. To complement deep contrast features, we also extract low-level features, which can effectively capture segment properties as well as color and texture contrasts between a region and the rest of the image. Low-level features are concatenated with deep contrast features to yield a hybrid deep and handcrafted feature vector. We show that training a random forest regressor over this hybrid feature vector can further boost the performance. Second, to enhance spatial coherence and better preserve the boundary of salient objects, a fully connected CRF model is integrated into our framework to perform pixelwise saliency refinement.

\begin{figure}[ht]
\begin{center}
   \includegraphics[width=0.45\textwidth]{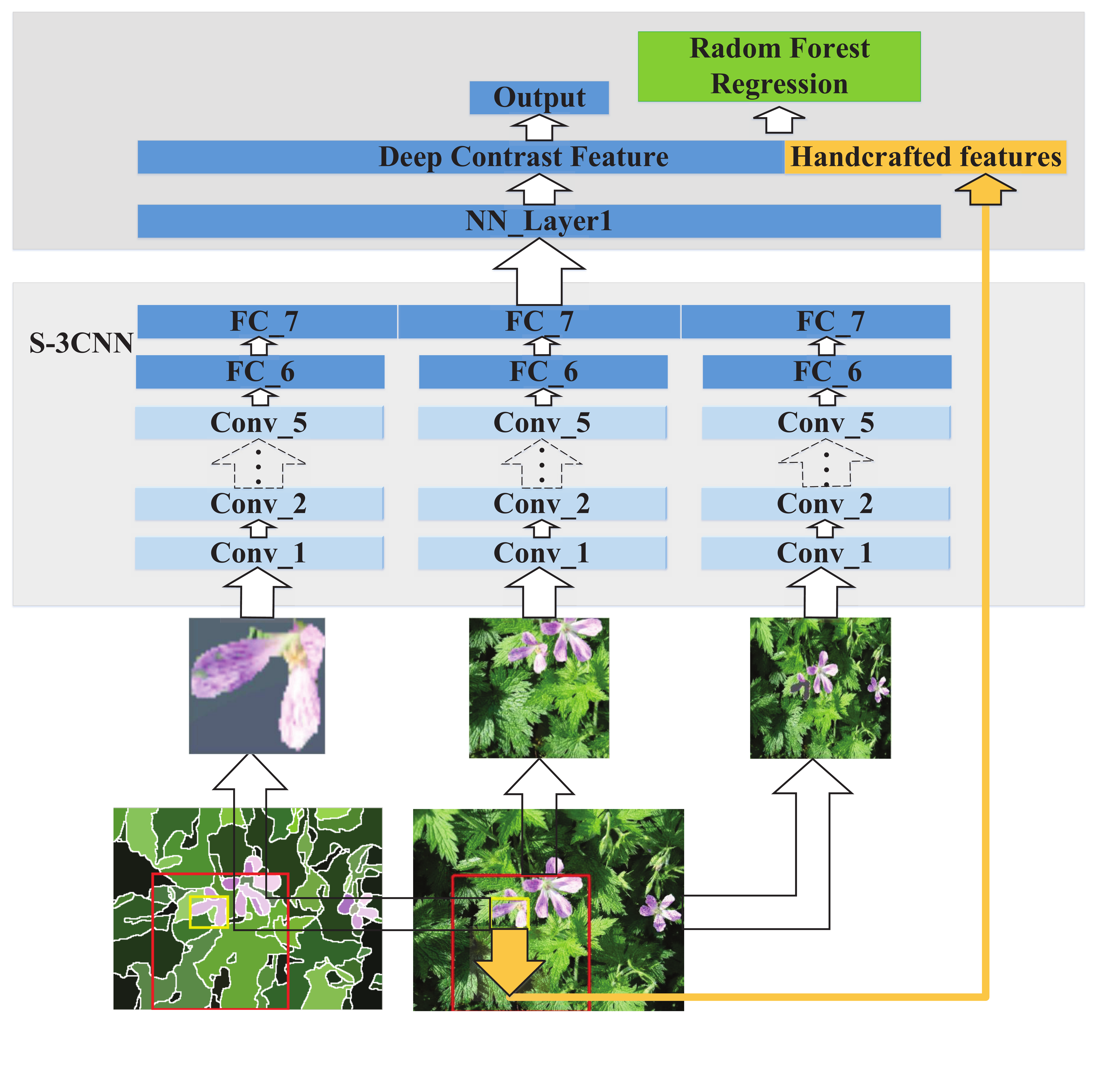}
\end{center}
   \caption{The architecture of our deep feature based visual saliency model.}
\label{fig:arch}
\end{figure}

%\IEEEpubidadjcol
\section{Saliency Inference with Deep Features}\label{sec:deep}
As shown in Fig. \ref{fig:arch}, the architecture of our deep feature based model for visual saliency consists of one output layer and two fully connected hidden layers on top of three deep convolutional neural networks. Our saliency model requires an input image to be decomposed into a set of nonoverlapping regions, each of which has almost uniform saliency values internally. The three deep CNNs are responsible for multiscale feature extraction. For each image region, they perform automatic feature extraction from three nested and increasingly larger rectangular windows, which are respectively the bounding box of the considered region, the bounding box of its immediate neighboring regions, and the entire image. The features extracted from the three CNNs are fed into the two fully connected layers, each of which has 300 neurons. The output of the second fully-connected layer is fed into the output layer, which performs logistic regression that infers the probability of a region being salient. When generating a saliency map for an input image, we run our trained saliency model repeatedly over every region of the image to produce a single saliency score for that region. This saliency score is further transferred to all pixels within that region. When the output of the penultimate layer is taken as a deep contrast feature, it can be concatenated with handcrafted low-level features to further boost saliency detection performance.

\subsection{Multiscale Feature Extraction}\label{sec:feature}
We extract multiscale features for each image region with a deep convolutional neural network originally trained over the ImageNet dataset~\cite{deng2009imagenet} and fine-tuned for object detection\cite{girshick2014rich} using Caffe~\cite{jia2014caffe}, an open source framework for CNN training and testing. The architecture of this CNN has eight layers including five convolutional layers and three fully-connected layers. Features are extracted from the output of the second last fully connected layer, which has 4096 neurons. Although this CNN was originally trained on datasets for visual recognition, automatically extracted CNN features turn out to be highly versatile and can be more effective than traditional handcrafted features on other visual computing tasks.

Since an image region may have an irregular shape while CNN features have to be extracted from a rectangular region, to make the CNN features only relevant to the pixels inside the region, as in \cite{girshick2014rich}, we define the rectangular region for CNN feature extraction to be the bounding box of the image region and fill the pixels outside the region but still inside its bounding box with the mean pixel values at the same locations across all ImageNet training images. These pixel values become zero after mean subtraction and do not have any impact on subsequent results.
We warp the region in the bounding box to a square with 227x227 pixels to make it compatible with the deep CNN trained for ImageNet. The warped RGB image region is then fed to the deep CNN and a 4096-dimensional feature vector is obtained by forward propagating a mean-subtracted input image region through all the convolutional layers and fully connected layers. We name this vector {\em feature A}.

Feature A itself does not include any information around the considered image region, thus is not able to tell whether the region is salient or not with respect to its neighborhood as well as the rest of the image. To include features from an area surrounding the considered region for understanding the amount of contrast in its neighborhood, we extract a second feature vector from a rectangular neighborhood, which is the bounding box of the considered region and its immediate neighboring regions. All the pixel values in this bounding box remain intact. Again, this rectangular neighborhood is fed to the deep CNN after being warped. We call the resulting vector from the CNN {\em feature B}.

As we know, a very important cue in saliency computation is the degree of (color and content) uniqueness of a region with respect to the rest of the image. The position of an image region in the entire image is another crucial cue. To meet these demands, we use the deep CNN to extract {\em feature C} from the entire rectangular image, where the considered region is masked with mean pixel values for indicating the position of the region. These three feature vectors obtained at different scales together define the features we adopt for saliency model training and testing. Since our final feature vector is the concatenation of three CNN feature vectors, we call it S-$3$CNN.

\subsection{Neural Network Training}\label{sec:train}
%The concatenated feature stated above still has not meet the requirement of our saliency detection, because the three parts of features were extracted in isolation, while indeed the relationship between these features is the criterion of a region being salient. Following this intuition, we create a neural network to dig the insight relationship of these three parts of features towards saliency estimation.

As discussed above, our proposed S-$3$CNN is a concatenation of three parts of deep features of 12288 dimmensions. On top of S-$3$CNN, we train a neural network with one output layer and two fully connected hidden layers. This network plays the role of a regressor that infers the saliency score of every image region from the multiscale CNN features extracted for the image region. It is well known that neural networks with fully connected hidden layers can be trained to reach a very high level of regression accuracy.

Concatenated multiscale CNN features are fed into this network, which is trained using a collection of training images and their labeled saliency maps, that have pixelwise binary saliency label. Before training, every training image is first decomposed into a set of regions. The saliency label of every image region is further estimated using pixelwise saliency labels. During the training stage, only those regions with 70\% or more pixels with the same saliency label are chosen as training samples, and their saliency score are set to either 1 or 0 respectively. During training, the output layer and the fully connected hidden layers together minimize the least-squares prediction errors accumulated over all regions from all training images.

\subsection{Deep Contrast Feature}\label{sec:dcf}
Note that the output of the penultimate layer of our neural network can be viewed as a fine-tuned feature vector for saliency detection. The final layer of our neural network essentially performs logistic regression on this fine-tuned feature, which effectively captures the contrast of a region with respect to its surrounding neighborhood at the semantic level. We name this feature Deep Contrast Feature (DCF) in the rest of this paper. Traditional regression techniques, such as support vector regression and boosted decision trees, can be trained on DCF to generate a saliency score for every image region. Nonetheless, we have found experimentally that this feature vector is highly discriminative and even simple logistic regression performed in the final layer of our neural network is sufficient to produce state-of-the-art performance on all visual saliency datasets. Since DCF reflects image semantics, we have further confirmed that DCF is complementary to handcrafted low-level features. In the following section, we show that training a random forest regressor over hybrid features including both DCF and some low-level regional features can further boost the performance.

\section{The Complete Algorithm}\label{sec:algo}
\subsection{Multi-Level Image Decomposition}\label{sec:region}
A variety of methods can be applied to decompose an image into nonoverlapping regions. Example methods include grids, region growing, and pixel clustering. Hierarchical image segmentation can generate regions at multiple scales to support the intuition that a semantic object at a coarser scale may be composed of multiple parts at a finer scale. In this paper, we applied the graph-based image segmentation\cite{felzenszwalb2004efficient} approach to compute \emph{M} levels of segmentation based on \emph{M} groups of segmentation parameters.
%, following the method described in \cite{jiang2013salient}. Since our approach train and performs saliency detection identically on each layer, we focus on a single layer of segmentation, denoting the specific layer of segmentation of $n$ segments as $S = \{ s_1, s_2, ..., s_n\}$.
Specifically, for an image \emph{I}, \emph{M} levels of image segmentations, $S = \{S_1, S_2, ..., S_M\} (|S_i|=N_i)$, are constructed from the finest to the coarsest scale. The regions at any level form a nonoverlapping image decomposition. %The hierarchical region merge algorithm in \cite{arbelaez2011contour} is applied to build a segmentation tree for the image. The initial set of regions are called superpixels.
%They are generated using the graph-based segmentation algorithm in \cite{felzenszwalb2004efficient}.
In our earlier version~\cite{li2015visual}, to generate a more accurate segmentation, region merger was prioritized by edge strength at boundary pixels shared by two adjacent regions and the edge strength was determined by an ultrametric contour map (UCM) proposed in \cite{arbelaez2011contour}. However, calculating UCM is time-consuming but does not clearly improve the accuracy of the final saliency map. %since multi-level segmentation and pixel-level CRF refinement method was applied for robustness purpose.
In this paper, we simply apply the graph-based segmentation algorithm in \cite{felzenszwalb2004efficient} to generate
15 levels of segmentations using different parameter settings.
The target number of regions at the finest and coarsest levels are controled to be around 300 and 20 respectively, and the number of regions at intermediate levels follows a geometric series. We train a unified model based on all the regions across these 15 levels of segmentations instead of a single model for each level of segmentation.

 %Regions with lower edge strength between them are merged earlier. The edge strength at a pixel is determined by a real-valued ultrametric contour map (UCM). In our experiments, we normalize the value of UCM into $[0,1]$ and generate 15 levels of segmentations with different edge strength thresholds. The edge strength threshold for level $i$ is adjusted such that the number of regions reaches a predefined target. The target number of regions at the finest and coarsest levels are set to 300 and 20 respectively, and the number of regions at intermediate levels follows a geometric series.

\begin{figure}[ht]
\begin{center}
   \includegraphics[width=0.45\textwidth]{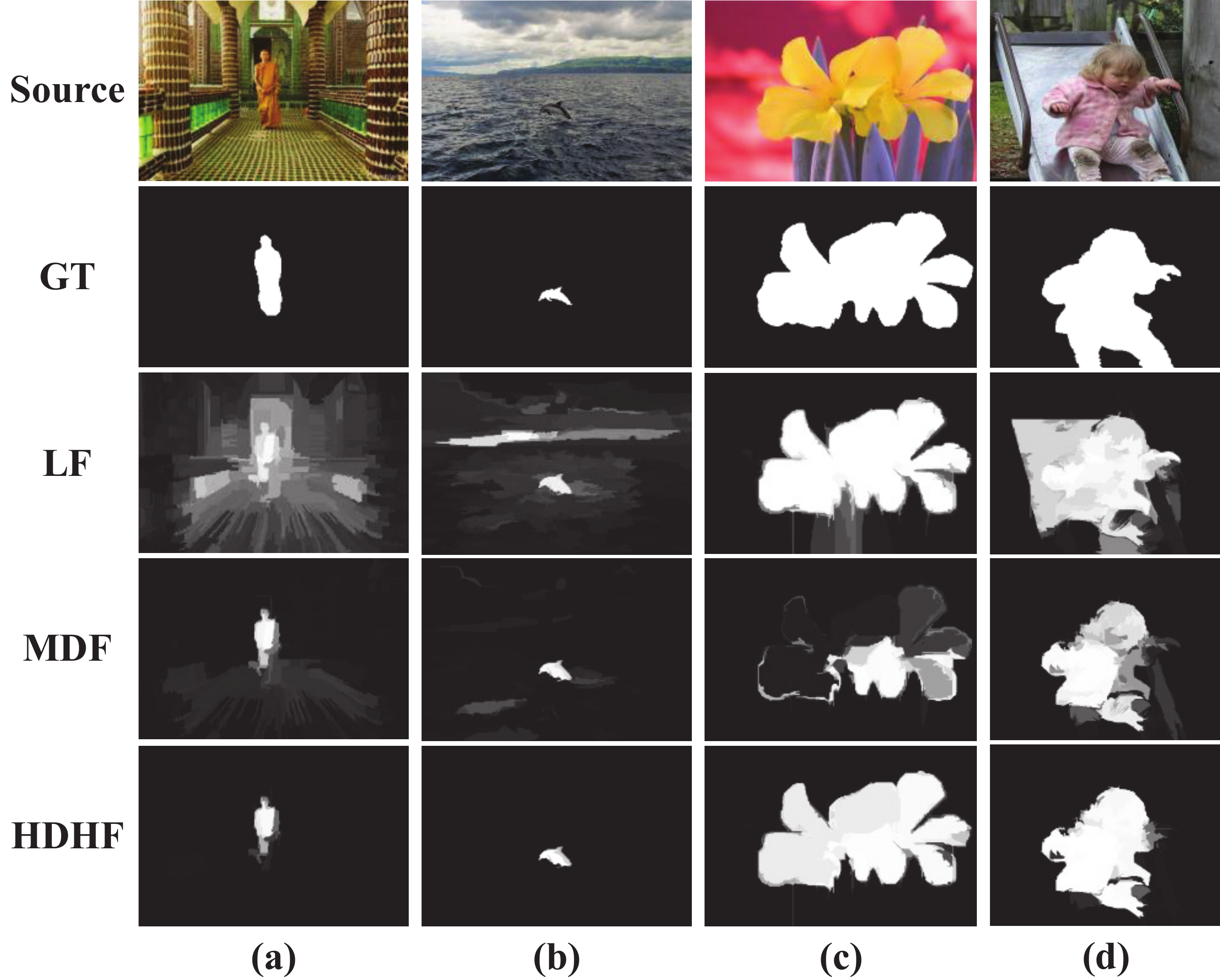}
\end{center}
   \caption{The integration of handcrafted low-level features with DCF. The ground truth (GT) is shown in the second row. LF denotes saliency maps generated using our defined handcrafted low-level feature. MDF refers to saliency maps generated using our multiscale deep feature. HDHF refers to saliency maps generated using hybrid deep and handcrafted feature. HDHF is consistently better than MDF and LF.}
\label{fig:mdf_and_hdcf}
\end{figure}

\subsection{HDHF: Hybrid Deep and Handcrafted Feature}
As discussed in Section~\ref{sec:dcf}, the initial saliency map from our trained neural network can be viewed as the result of logistic regression on DCF. As shown in Fig.~\ref{fig:mdf_and_hdcf}, DCF is especially adept at detecting salient regions in images with low contrast and complex background as long as there exists semantic distinction against their surrounding neighborhoods (Fig.~\ref{fig:mdf_and_hdcf}~a\&b). However, since DCF is derived from multi-scale CNN features that are focused on image semantics, it may not contain sufficient information about the contrast in low-level attributes. For example, as shown in Fig.~\ref{fig:mdf_and_hdcf}~c, when regions are salient due to contrast in low-level attributes (e.g. color and texture), DCF tends to perform worse than those methods based on handcrafted low-level features. And there are many examples where neither deep features nor handcrafted low-level features alone are good enough to generate accurate saliency maps (e.g. Fig.~\ref{fig:mdf_and_hdcf}~a,c\&d)). To overcome this deficiency, we propose a small set of complementary low-level features to compensate DCF.
%Different from~\cite{jiang2013salient},  where backgroundness descriptor and center bias are proved to be not applicable for challenging dataset such as HKU-IS, we use the most common priors, such as colour and texture contrast and a modified backgroundness description as well as some segment properties.
%\subsubsection{Handcrafted Low-level Feature Discription}

Given an image, we first generate an initial saliency map $SM^{init}$ using multiscale deep features. We define a pseudo background region $B$ as the set of pixels within 30 pixels from the image borders and having an initial saliency value $SM^{init} < 0.1$. We compute low-level features for the entire image, the pseudo background region, and every region in every level of image segmentation. Such low-level features includes both color and texture features. Color features include RGB, LAB and HSV histograms as well as their corresponding average values. Texture features include the histogram of maximum responses of $LM$ filters as well as the histogram of $LBP$ features.

On the basis of these low-level features, for each region $R$ in each level of segmentation, we extract a 39-dimensional low-level feature descriptor including both contrast features and segment properties. The contrast features include the contrast between the low-level features of $R$ and their corresponding features of the pseudo background $B$ as well as the contrast between the low-level features of $R$ and their counterparts for the entire image. We adopt the $\chi^2$ distance as the contrast between two histograms and the absolute difference as the contrast between two scalar features. Segment properties include the variance of various color and texture features as well as geometric properties including the perimeter and area of the segment. Note that the geometric properties are normalized with respect to the overall image size. The details of all handcrafted low-level features are given in Table~\ref{tab:feature}. We normalize the $L_2$ norm of both our proposed 300-dimensional DCF and this handcrafted low-level feature descriptor before concatenating them into a 339-dimensional hybrid feature vector, called hybrid deep and handcrafted feature (HDHF).

\begin{table*}[]
\centering
\resizebox{0.9\textwidth}{!}
{
\renewcommand{\arraystretch}{2.0}
\begin{tabular}{|c|c|c|c|c|c|c|c|}
\hline
\multicolumn{4}{|c|}{Contrast Descriptors (Color and Texture)}                                                                                       & \multicolumn{4}{c|}{Segment Properties}                                                        \\ \hline
Notation             & Features                                       & Definition                                                             & Dim & Notation       & Features                & Definition                                  & Dim \\ \hline
$c_1 \sim c_6$       & Difference between Average RGB Values               & $|R^{rgb}-B^{rgb}|$, $|R^{rgb}-I^{rgb}|$                               & 6   & $s_1 \sim s_3$ & Variances of RGB values & $var_{R}^{r}$, $var_{R}^{g}$, $var_{R}^{b}$ & 3   \\ \hline
$c_7 \sim c_8$       & $\chi^2$ distance between RGB Histograms             & $\chi^2(h_{rgb}^{R}, h_{rgb}^{B})$, $\chi^2(h_{rgb}^{R}, h_{rgb}^{I})$ & 2   & $s_4 \sim s_6$ & Variances of LAB values & $var_{R}^{l}$, $var_{R}^{a}$, $var_{R}^{b}$ & 3   \\ \hline
$c_9 \sim c_{14}$    & Difference between Average LAB Values               & $|R^{lab}-B^{lab}|$, $|R^{lab}-I^{lab}|$                               & 6   & $s_7 \sim s_9$ & Variance of HSV values & $var_{R}^{h}$, $var_{R}^{s}$, $var_{R}^{v}$ & 3   \\ \hline
$c_{15} \sim c_{16}$ & $\chi^2$ distance between LAB Histograms             & $\chi^2(h_{lab}^{R}, h_{lab}^{B})$, $\chi^2(h_{lab}^{R}, h_{lab}^{I})$ & 2   & $s_{10}$       & Normalized perimeter    & $Perimeter(R)$                                & 1   \\ \hline
$c_{17} \sim c_{22}$ & Difference between Average HSV Values               & $|R^{hsv}-B^{hsv}|$, $|R^{hsv}-I^{hsv}|$                               & 6   & $s_{11}$       & Normalized area         & $Area(R)$                                     & 1   \\ \hline
$c_{23} \sim c_{24}$ & $\chi^2$ distance between HSV Histograms             & $\chi^2(h_{hsv}^{R}, h_{hsv}^{B})$,$\chi^2(h_{hsv}^{R},  h_{hsv}^{I})$ & 2   &                &                         &                                             &     \\ \hline
$c_{25} \sim c_{26}$ & $\chi^2$ distance b.w. Max response LM Histograms & $\chi^2(h_{LM}^{R}, h_{LM}^{B})$, $\chi^2(h_{LM}^{R}, h_{LM}^{I})$     & 2   &                &                         &                                             &     \\ \hline
$c_{27} \sim c_{28}$ & $\chi^2$ distance between LBP Histograms             & $\chi^2(h_{LBP}^{R}, h_{LBP}^{B})$, $\chi^2(h_{LBP}^{R}, h_{LBP}^{I})$ & 2   &                &                         &                                             &     \\ \hline
\end{tabular}
}
\caption{A detailed description of handcrafted low-level features. $R$ denotes an image segment, $B$ refers to the pseudo background region, and $I$ denotes the entire image.}
\label{tab:feature}
\end{table*}

\subsection{Training Saliency Regressor over HDHF}
To demonstrate the effectiveness of HDHF, we train a random forest regressor using hybrid deep and handcrafted features. Each training sample corresponds to a region with a 339-dimensional HDHF. As done for neural network training in Section~\ref{sec:train}, only those regions with 70\% or more pixels with the same saliency label are chosen as training samples, and their saliency scores are set to either 1 or 0 accordingly. Learning a random forest based model can automatically integrate low-level and high-level features, and map every HDHF to a saliency score. We also train another random forest model base only on 39 dimensional low-level features for comparison. As shown in Fig.~\ref{fig:mdf_and_hdcf} and the quantitative results in Section~\ref{sec:per_hdcf}, HDHF based saliency maps are consistently better than those based on DCF or handcrafted features only.

\subsection{Saliency Map Fusion}\label{sec:fusion}
Given the regions in an image decomposition, we can generate an initial saliency map either with the neural network model or the HDHF-based random forest regressor. Given $M$ levels of segmentations, we obtain $M$ saliency maps, $\{A^{(1)}, A^{(2)}, ..., A^{(M)}\}$, interpreting salient parts of the input image at various granularity. We aim to further fuse them together to obtain an aggregated saliency map.
To this end, we take a simple approach by assuming the aggregated saliency map is a linear combination of the maps at individual segmentation levels, and learn the weights in the linear combination by running a least-squares estimator over a validation dataset. %indexed with $I_v$.
Thus, our aggregated saliency map $A$ is formulated as follows,
{\small
\begin{equation}
\begin{split}
&A = \sum_{k = 1}^{M} \alpha_k A^{(k)} \\
\text{s.t. }  \{\alpha_k\}_{k=1}^{M} &= \argmin_{\alpha_1, \alpha_2, ..., \alpha_M} \sum_{i\in I_v}\left\|A_i - \sum_{k} \alpha_k A_i^{(k)}\right\|^2_F.
%&A = \sum_{i = 1}^{M} (\alpha_i A^{(i)}) \\
% \text{s.t. }  \{\alpha_i\}_{i=1}^{M} &= \argmin_{\alpha_1, \alpha_2, ..., \alpha_M}||A - \sum_{i} \alpha_i A^{(i)}||_F.
\end{split}
\end{equation}
}
where $I_v$ stands for the set of indices of the images in the validation dataset.

Note that there are many options for saliency fusion. For example, a conditional random field (CRF) framework has been adopted in \cite{mai2013saliency} to aggregate multiple saliency maps from different methods. %which can appropriately consider the performance gaps among individual methods and generate a saliency map better than any individual one. Since the performance gap on each layer saliency map is not as large as those from various methods,
Nevertheless, we have found that, in our context, a linear combination of all saliency maps can already serve our purposes well and is capable of producing aggregated maps with a quality comparable to those obtained from more complicated techniques.

\begin{figure}[ht]
\begin{center}
   \includegraphics[width=0.45\textwidth]{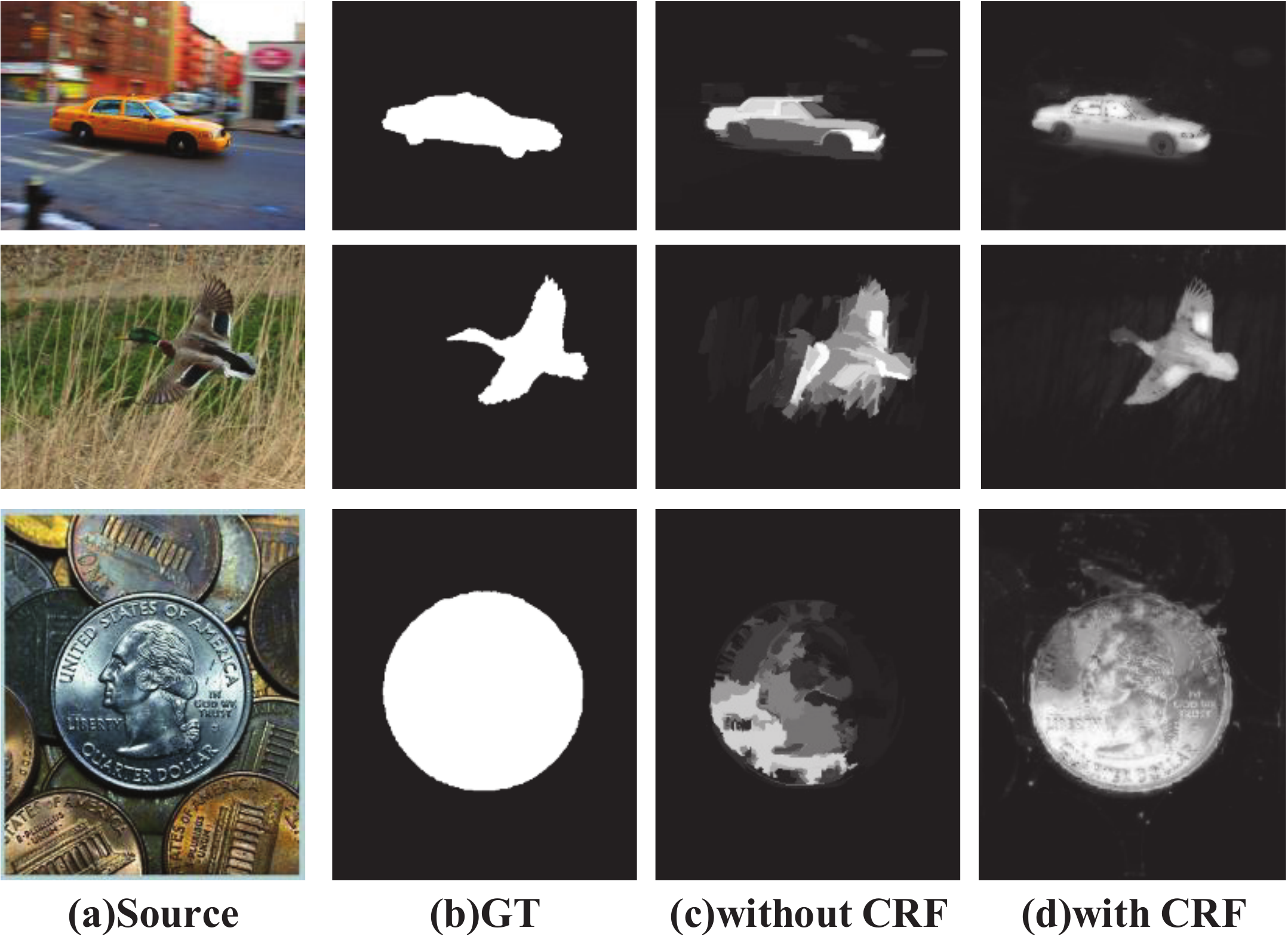}
\end{center}
   \caption{Comparison of saliency detection results with and without CRF.}
\label{fig:crf_effect}
\end{figure}

\subsection{Spatial Coherence Based on CRF}\label{sec:coherence}
Due to the fact that image segmentation is imperfect and our model assigns saliency scores to individual segments, noisy scores inevitably appear in the above aggregated saliency map. To enhance spatial coherence, we perform pixelwise saliency refinement using the fully connected CRF model in \cite{krahenbuhl2012efficient}. This model solves a binary pixel labeling problem, and employs the following energy function,
\begin{equation}
E\left( L \right) = -\sum_{i}\log P\left(l_i\right)+\sum_{i,j}\theta_{ij}\left(l_i, l_j\right),
\end{equation}
where $L$ represents a binary label (salient or not salient) assignment for all pixels. $P(l_i)$ is the probability of pixel $x_i$ having label $l_i$, which indicates the likelihood of pixel $x_i$ being salient. Initially, $P(1)=S_i$ and $P(0)=1-S_i$, where $S_i$ is the saliency score at pixel $x_i$ from the above aggregated saliency map $A$. $\theta_{ij}\left(l_i,l_j\right)$ is a pairwise potential and defined as follows,
\begin{equation}
\begin{split}
\theta_{ij}=\mu\left(l_i,l_j\right)\Bigg[ \omega_1\exp\Bigg(-\frac{\left \|p_i-p_j  \right \|^2}{2\sigma_\alpha^2}-\frac{\left \|I_i-I_j \right \|^2}{2\sigma_\beta^2}\Bigg) +\\ \omega_2\exp\left(-\frac{\left \|p_i-p_j \right\|^2}{2\sigma_\gamma^2}\right)\Bigg],
\end{split}
\end{equation}
where $\mu\left(l_i,l_j\right) = 1$ if $l_i \neq l_j$, and zero otherwise. $\theta_{ij}$ involves two kernels. The first bilateral kernel depends on both pixel positions (denoted as $p$) and colors (denoted as $I$), suggesting nearby pixels with similar colors to be assigned similar saliency scores. The degrees of color similarity and pixel closeness are controlled by two parameters, $\sigma_\alpha$ and $\sigma_\beta$, respectively. The second kernel only depends on pixel position and aims at removing small isolated regions. The ``scale'' of the Gaussian kernel is controlled by $\sigma_\gamma$. The parameters are determined through cross validation using the validation set of MSRA-B dataset in our experiment, as in~\cite{krahenbuhl2012efficient}.

Energy minimization is based on a mean field approximation to the CRF distribution and high-dimensional filtering can be utilized to speed up computation. In this paper, we use the publicly available implementation of \cite{krahenbuhl2012efficient} to minimize the energy, and it takes less than 0.5 second on an image with $300\times 400$ pixels.  At the end of energy minimization, we generate a saliency map using the posterior probability of each pixel being salient. Note that features other than color can be used in the first term to boost performance (e.g. contour information was used in an earlier version of this paper~\cite{li2015visual}). Currently, we only use color for the sake of efficiency and find it sufficient for enhancing spatial coherence and removing noisy saliency scores in the aggregated saliency map due to imperfect segmentation. The result is an enhanced saliency map. As shown in Fig.~\ref{fig:crf_effect}, our initial saliency maps in general look fragmented and the boundaries of salient objects are not well preserved. The application of the CRF model can not only give rise to smoother results with pixelwise accuracy but also better preserve the boundaries of salient objects.
A quantitative study of the effectiveness of the CRF model can be found in Section~\ref{sec:spatial_coherence}.

\section{A New Dataset}\label{sec:data}
%At present, the pixelwise ground truth annotation~\cite{jiang2013salient} of the MSRA-B dataset~\cite{liu2011learning} is the only large dataset that is suitable for training a deep neural network. Nevertheless, this benchmark becomes less challenging once a center prior and a boundary prior~\cite{jiang2013salient,zhu2014saliency} have been imposed since most images in the dataset contain only one connected salient region and 98\% of the pixels in the border area belongs to the background~\cite{jiang2013salient}.

% We have constructed a more challenging dataset to facilitate the research and evaluation of visual saliency models. our new saliency dataset, called HKU-IS, contains 4447 images with high-quality pixelwise annotations. It is more challenging and unbiased compared with the most often used dataset~(e.g. MSRA-B~\cite{liu2011learning}) and of much less ambiguity than DUT-OMRON~\cite{yang2013saliency}. Please refer to the preliminary version of this paper~\cite{li2015visual} on the detailed construction of the new dataset.

We have constructed a more challenging dataset to facilitate the research and evaluation of visual saliency models. To build the dataset, we initially collected 7320 images. These images were chosen by following at least one of the following criteria:
\begin{enumerate}
\item there are multiple disconnected salient objects;\vspace{-1mm}
\item at least one of the salient objects touches the image boundary;\vspace{-1mm}
\item the background is complex;
\item the color contrast (the minimum Chi-square distance between the color histograms of any salient object and its surrounding regions) is less than 0.7.
\end{enumerate}
To reduce label inconsistency, we asked three people to annotate salient objects in all 7320 images individually using a custom designed interactive segmentation tool. On average, each person takes 1-2 minutes to annotate one image. The annotation stage spanned over three months.

Let $A^p = \{a_x^{(p)}\}$ be the binary saliency mask labeled by the $p$-th user. And $a_x^{(p)} = 1$ if pixel $x$ is labeled as salient and $a_x^{(p)} = 0$ otherwise. We define label consistency as the ratio between the number of pixels labeled as salient by all three people and the number of pixels labeled as salient by at least one of the people. It is formulated as
\begin{equation}
C = \frac{\sum_x \left( \prod_{p=1}^{3}a_{x}^{(p)} \right)}{\sum_x \mathbf{1}\left( \sum_{p=1}^{3}a_{x}^{(p)} \neq 0 \right)}.
\end{equation}

We excluded those images with label consistency $C < 0.9$, and 4447 images remained. For each image that passed the label consistency test, we generated a ground truth saliency map from the annotations of three people. The pixelwise saliency label in the ground truth saliency map, $G = \{g_x|g_x \in \{0, 1\}\}$, is determined according to the majority label among the three people as follows,
\begin{equation}
g_x = \mathbf{1}\left(\sum_{p = 1}^{3}a_x^{(p)} \geq 2 \right).
\end{equation}

At the end, our new saliency dataset, called HKU-IS, contains 4447 images with high-quality pixelwise annotations. It is more challenging and unbiased compared with the most often used dataset~(e.g. MSRA-B~\cite{liu2011learning}). % and of much less ambiguity than DUT-OMRON~\cite{yang2013saliency}.

%All the images in HKU-IS satisfy at least one of the above three criteria while 2888 (out of 5000) images in the MSRA dataset do not satisfy any of these criteria. In summary, 50.34\% images in HKU-IS have multiple disconnected salient objects while this number is only 6.24\% for the MSRA dataset; 21\% images in HKU-IS have salient objects touching the image boundary while this number is 13\% for the MSRA dataset; and the mean color contrast of HKU-IS is 0.69 while that of the MSRA dataset is 0.78.

\section{Experimental Results}\label{sec:experiment}
\subsection{Dataset}
We have evaluated the performance of our method on several public benchmarks for salient object detection as well as on our own dataset.

{\flushleft \textbf{MSRA-B}}\cite{liu2011learning}. This dataset has 5000 images, and is widely used for salient object detection. Most of the images contain only one salient object. Pixelwise annotation was provided by \cite{jiang2013salient}.
\vspace{-0mm}

{\flushleft \textbf{DUT-OMRON}}\cite{yang2013saliency}. This large dataset contains 5168 natural images. Both bounding boxes and pixelwise salient object annotations are provided. We have noticed that many saliency annotations in this dataset may be controversial among different human observers. As a result, none of the existing saliency models has achieved a high accuracy on this dataset.
%{\flushleft \textbf{SED}}\cite{alpert2007image}. It contains two subsets: SED1 and SED2. SED1 has 100 images each containing only one salient object while SED2 has 100 images each containing two salient objects.
\vspace{-0mm}
{\flushleft \textbf{SOD}}\cite{MartinFTM01}. This dataset has 300 images, and it was originally designed for image segmentation. Pixelwise annotation of salient objects in this dataset was generated by \cite{jiang2013salient}. This dataset is very challenging since many images contain multiple salient objects either with low contrast or overlapping with the image boundary.
%\vspace{-0mm}
%{\flushleft \textbf{iCoSeg}}\cite{batra2010icoseg}. This dataset was designed for co-segmentation. It contains 643 images with pixelwise annotation. Each image may contain one or multiple salient objects.
\vspace{-0mm}
{\flushleft \textbf{PASCAL-S}}\cite{li2014secrets}. This dataset was built using the validation set of the PASCAL VOC 2010 segmentation challenge. It contains 850 images with pixelwise salient object annotation. The groundtruth saliency masks were labeled by 12 subjects. We threshold the masks at 0.5 to obtain binary masks as suggested in \cite{li2014secrets}.

{\flushleft \textbf{ECSSD}}\cite{yan2013hierarchical}. This dataset contains 1,000 structurally complex images acquired from the Internet with pixelwise groundtruth masks.

{\flushleft \textbf{HKU-IS}}. Our new dataset contains 4447 images with pixelwise annotation of salient objects.

To save space, the performance on the SED~\cite{alpert2007image} and ICOSEG~\cite{batra2010icoseg} datasets is no longer reported since these datasets are not challenging and not widely used. Readers can refer to an earlier version of our paper~\cite{li2015visual} for performance comparisons on these two datasets. To facilitate a fair comparison with other methods, we divided the MSRA dataset into three parts as in~\cite{jiang2013salient}, 2500 for training, 500 for validation and the remaining 2000 images for testing. To test the adaptability of trained saliency models to other different datasets, we use the models trained on the MSRA-B dataset and test them over all other datasets.

As discussed in the previous sections, we generate two sets of saliency results using our proposed saliency models. To evaluate the effectiveness of multiscale deep features, we construct the first set of saliency maps from the output of the neural network model aggregated with multi-level fusion and further enhanced using the CRF model. To demonstrate the complementariness between DCF and handcrafted low-level features, we generate the second set of saliency maps from the random forest regressor using HDHF, also aggregated with multi-level fusion and enhanced using the CRF model. We refer to the first set of saliency maps as MDF, and the second set HDHF in the rest of this paper. When conducting an ablation study, we investigate the contribution of each component to the accuracy of MDF, and we show the overall performance of both MDF and HDHF when comparing them with other state-of-the-art methods.

\begin{figure}[ht]
    \centerline{
    \includegraphics[width = 0.248\textwidth]{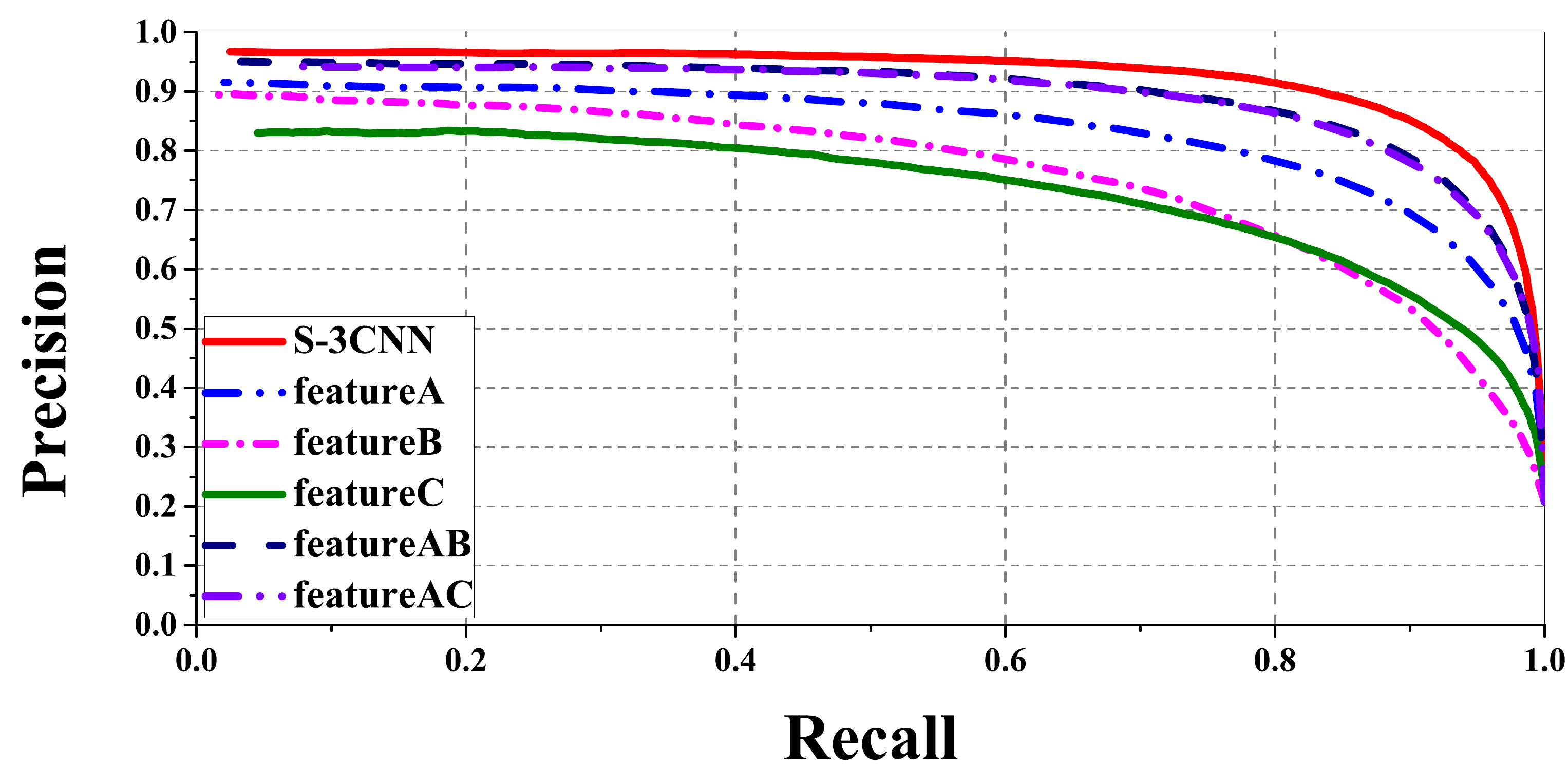}\hfill
    \includegraphics[width = 0.248\textwidth]{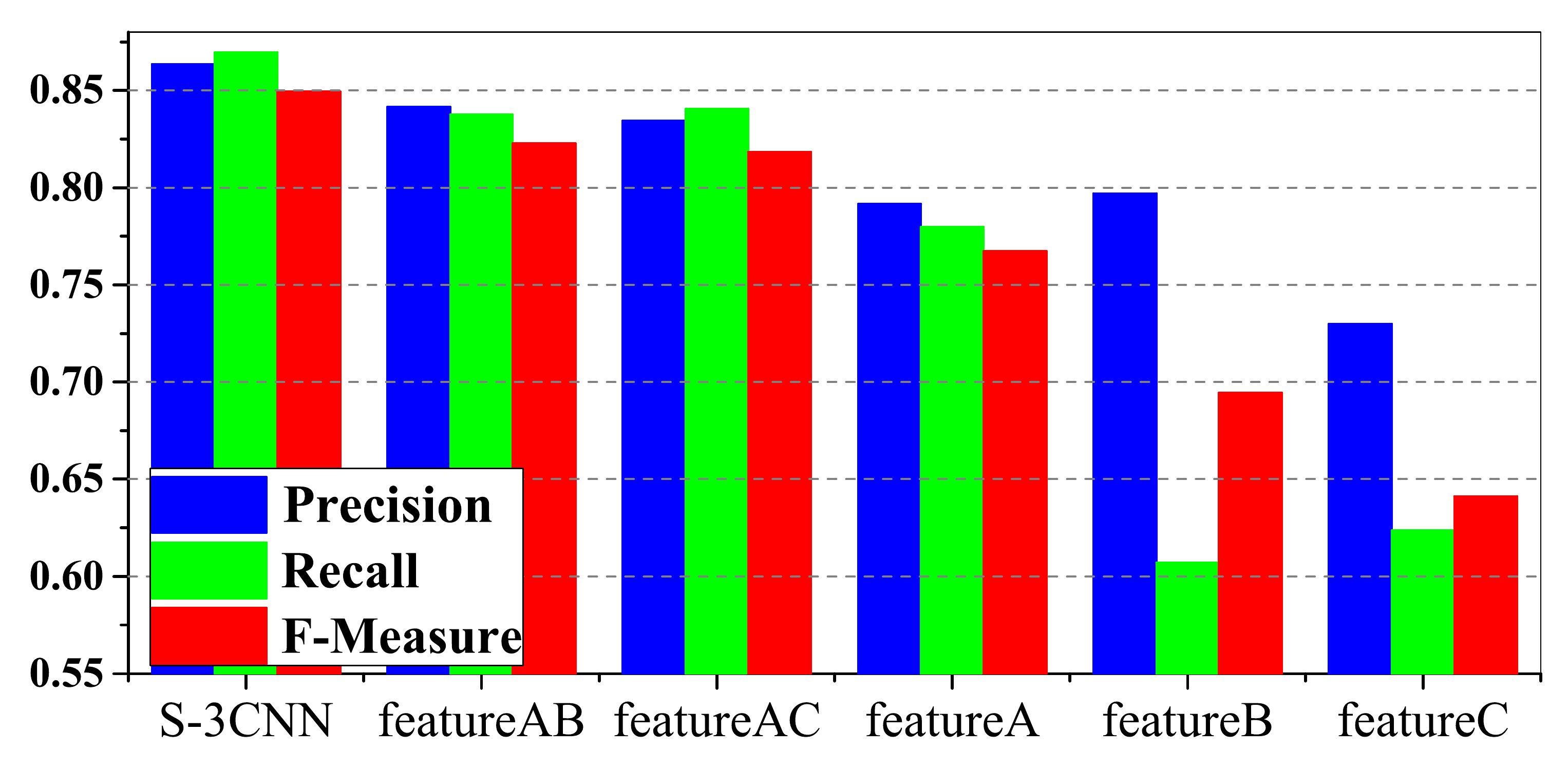}\hfill
  }\vspace{-0mm}
    %\centerline{\hfill (a) \hfill\hfill  (b) \hfill}\vspace{-1mm}
    \caption{The effectiveness of our S-3CNN feature. The left figure shows the precision-recall curves of models trained on MSRA-B using different components of S-3CNN, while the right figure shows the corresponding precision, recall and F-measure using an adaptive threshold.}
\label{fig:analysis_on_s3cnn}
\end{figure}

\begin{figure}[ht]
    \centerline{
    \includegraphics[width = 0.248\textwidth]{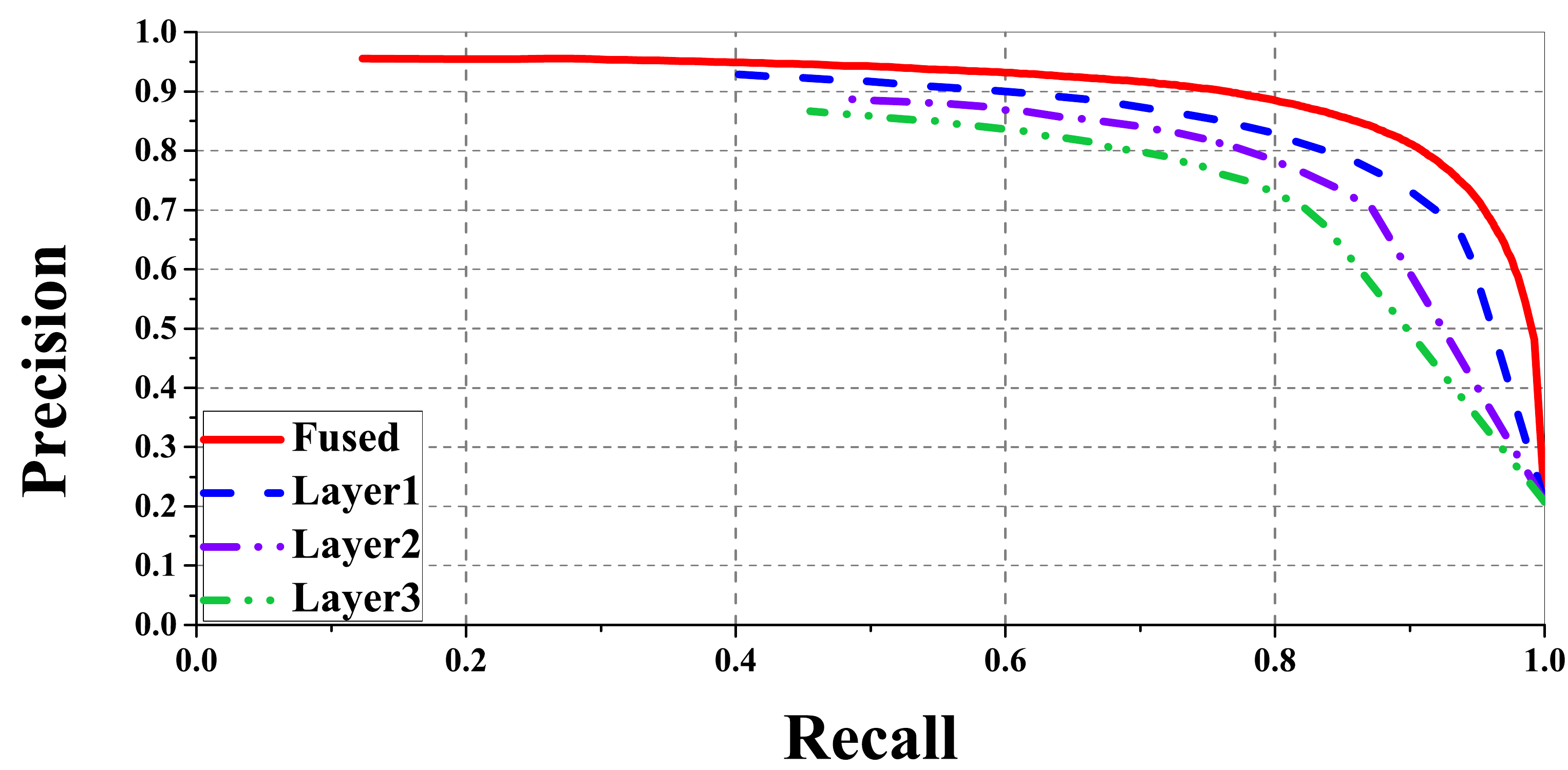}\hfill
    \includegraphics[width = 0.248\textwidth]{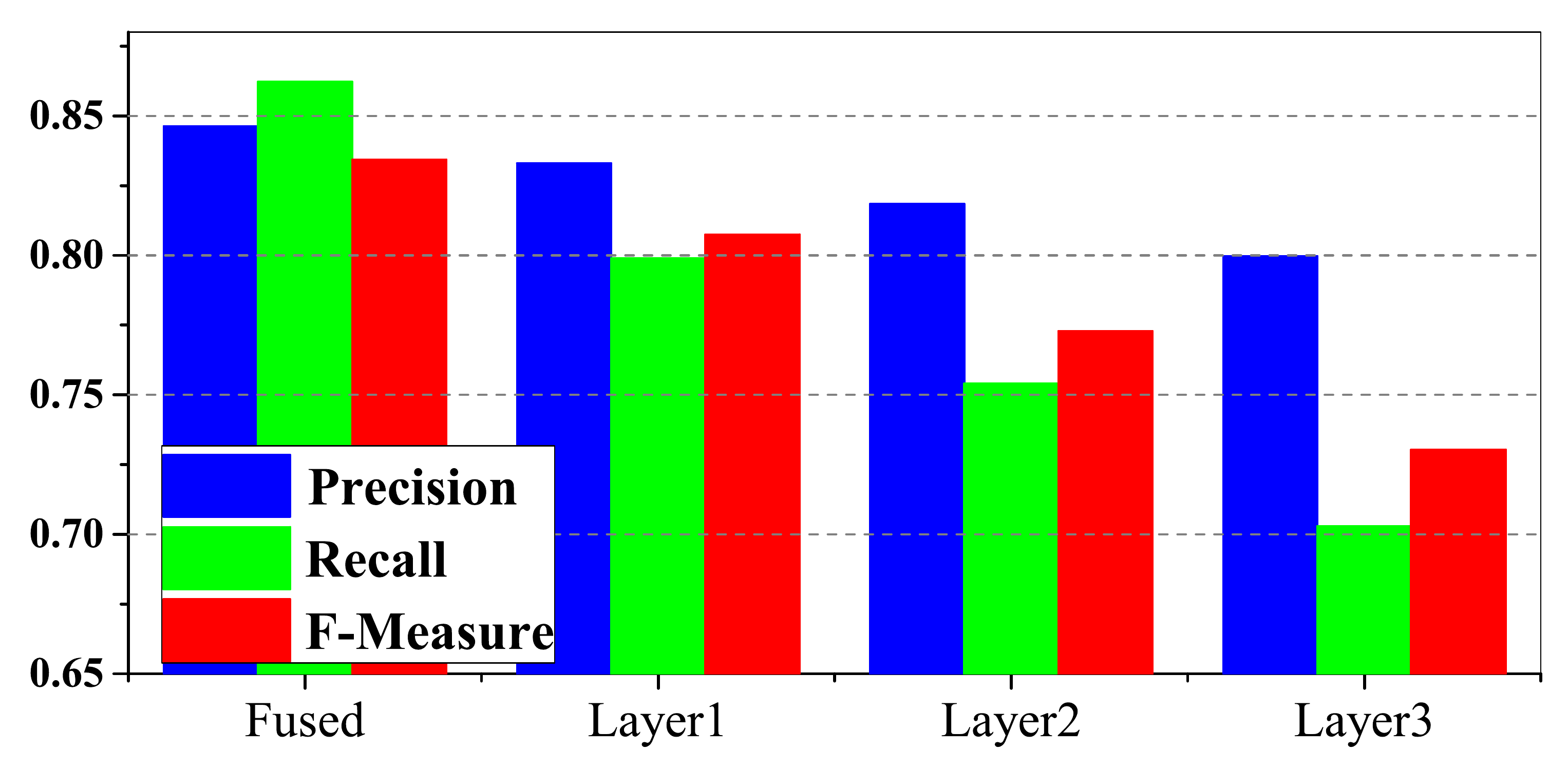}\hfill
  }\vspace{-0mm}
    %\centerline{\hfill (a) \hfill\hfill (b) \hfill\hfill (c) \hfill\hfill (d) \hfill}\vspace{-1mm}
    \caption{The effectiveness of multilevel fusion.``Layer1", ``Layer2" and ``Layer3" refer to the three segmentation levels that have the highest single-level saliency detection performance. The left figure shows the precision-recall curves while the right figure shows the corresponding precision, recall and F-measure using an adaptive threshold.}
\label{fig:analysis_on_multilayer}
\end{figure}

\begin{figure}[ht]
    \centerline{
    \includegraphics[width = 0.248\textwidth]{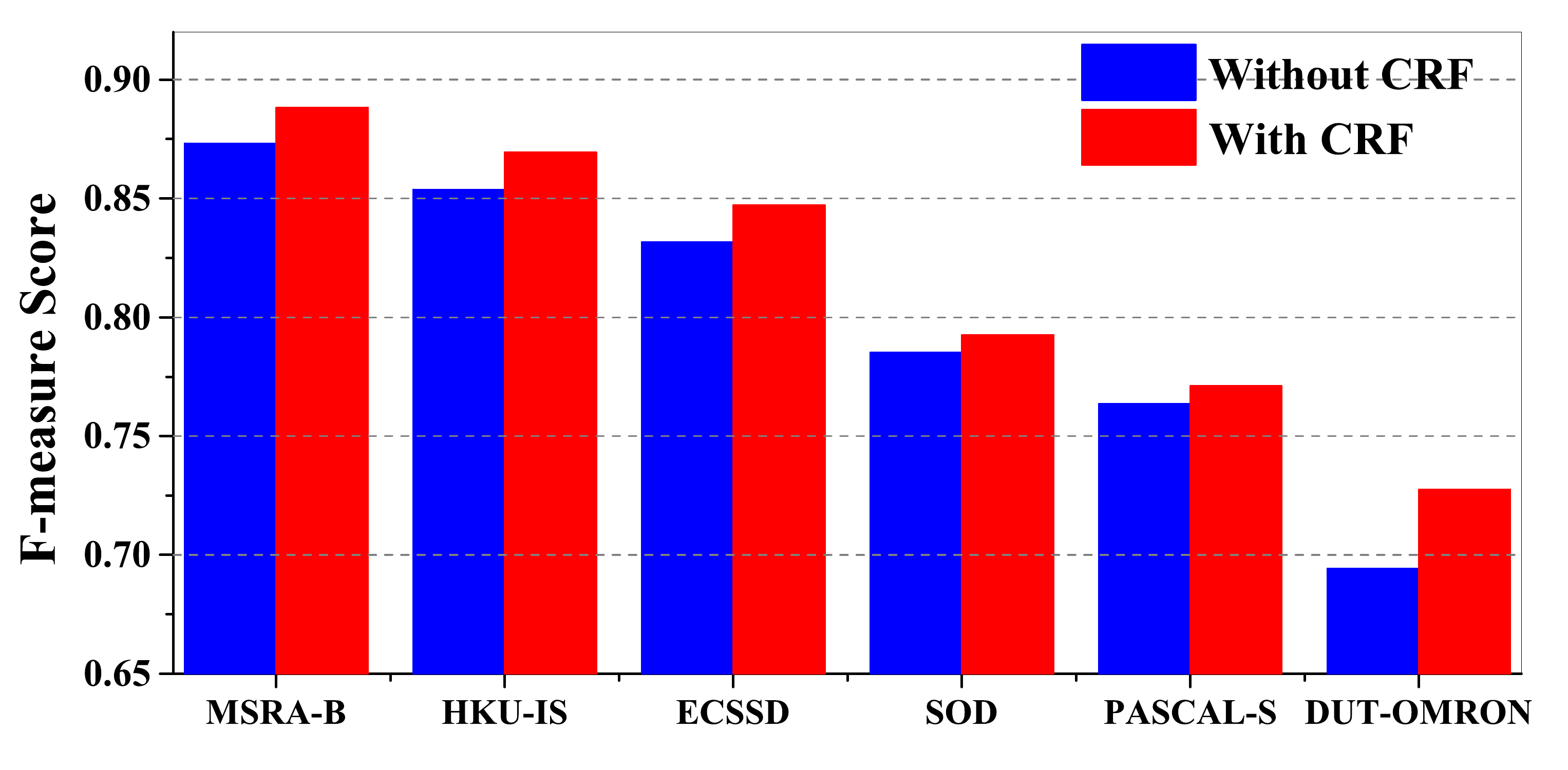}\hfill
    \includegraphics[width = 0.248\textwidth]{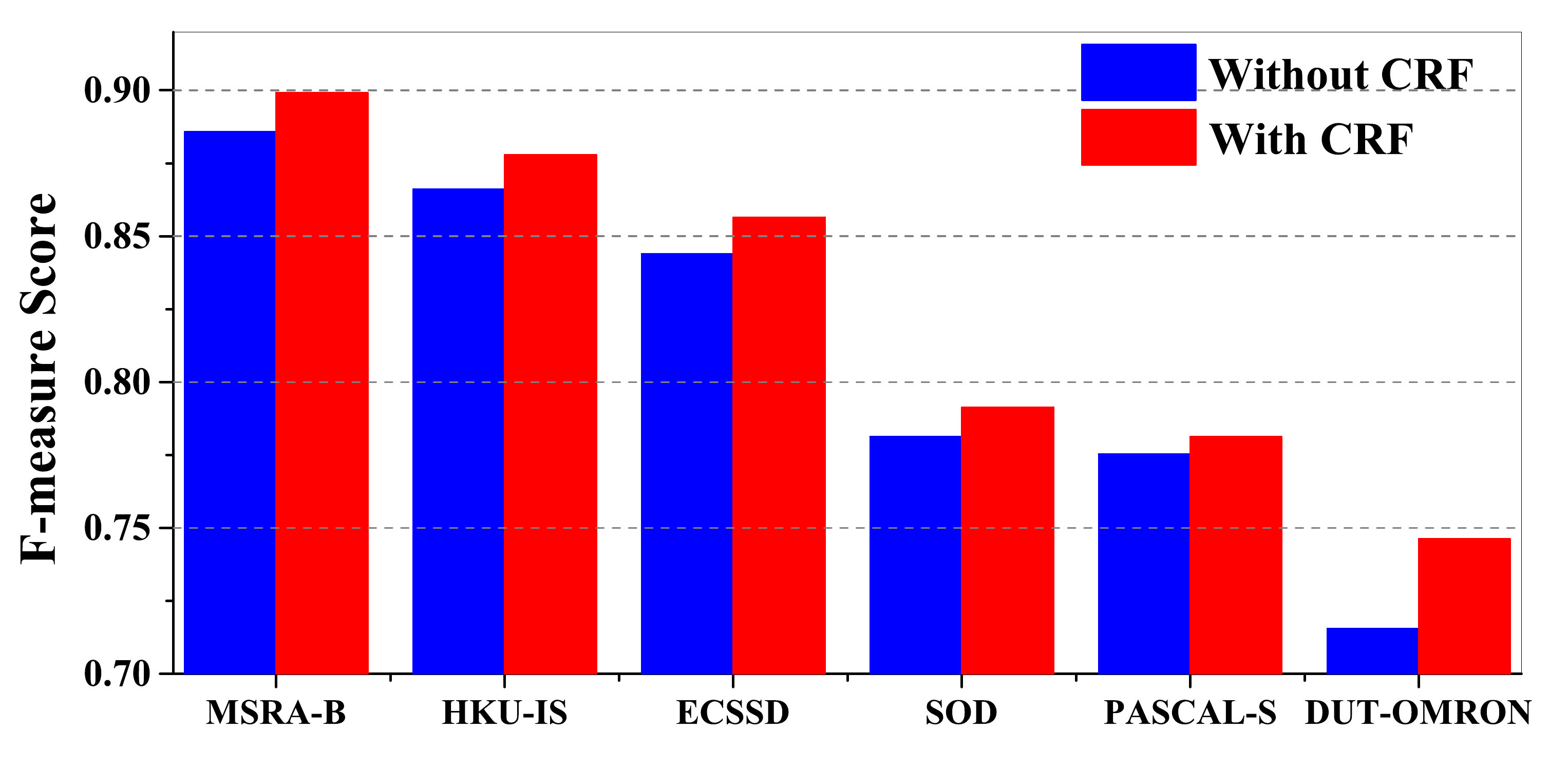}
  }\vspace{-0mm}
    \centerline{\hfill (a) \hfill\hfill (b) \hfill}\vspace{-1mm}
    \caption{The effectiveness of CRF-based spatial coherence. (a) Maximum F-measure of our MDF-based model achieved with and without CRF on six saliency detection datasets. (b)Maximum F-measure of our HDHF-based model achieved with and without CRF on the same datasets.}
\label{fig:spatial_coherence}
\end{figure}

\subsection{Implementation Details}
We train our saliency models using the training set of the MSRA-B dataset and test them over all other datasets. The training set contains 2500 images. We perform 15 levels of image segmentation and extract around 800 segments across all levels from each image. The S-$3$CNN feature vector extracted from each segment forms one training sample, and we have 1.9 million training samples in total. Though the dimension of S-$3$CNN and HDHF are larger than 12 thousand, the number of our training samples is large enough to train an accurate model free from overfitting. We have implemented our proposed framework in Caffe~\cite{jia2014caffe}. More specifically, to train our three-layer perceptron network, the learning rate is set to 0.2 while the momentum parameter is set to 0.5. We use the hyperbolic tangent function as the activation function in the hidden layers and the sigmoid function in the output layer. When jointly fine-tune deep CNN model with our proposed three-layer MLP, the learning rate of the initial deep CNN model is set to 0.0001. We cross-validate the parameters in the fully connected CRF according to~\cite{krahenbuhl2012efficient} on the validation set and the final values of $w_1$, $w_2$, $\sigma_\alpha$, $\sigma_\beta$, and $\sigma_\gamma$ are $3.0$, $5.0$, $3.0$, $50.0$ and $3.0$ respectively.

\subsection{Evaluation Criteria}
Following \cite{borji2012salient,ChengPAMI}, we first use standard precision-recall~(PR) and receiver operating characteristic~(ROC) curves to evaluate the performance of our method.
%Precision measures the percentage of correctly assigned salient pixels while recall calculates the percentage of salient pixels detected.
A continuous saliency map can be converted to a binary mask using a threshold, resulting in a pair of precision and recall values when the binary mask is compared against the ground truth. A PR curve is then obtained by varying the threshold from $0$ to $1$. The curves are averaged over each dataset. The ROC curve can be conveniently generated according to the true positive rates and false positive rates obtained during the calculation of the PR curve. The AUC~(Area Under ROC Curve) score is also reported given the ROC curve.

Second, since high precision and high recall are both desired in many applications, we compute the F-measure\cite{achanta2009frequency} as
\begin{equation}
 F_{\beta} = \frac{(1+\beta^2)\cdot Precision \cdot Recall}{\beta^2\cdot Precision + Recall},
\end{equation}
where $\beta^2$ is set to 0.3 to weigh precision more than recall as suggested in \cite{achanta2009frequency}. We report the maximum F-measure score among all pairs of precision and recall values. We also report the performance once every saliency map is binarized with an image-dependent threshold proposed by \cite{achanta2009frequency}. This adaptive threshold is determined to be twice the mean saliency of the image:
\begin{equation}
 T_a = \frac{2}{W \times H} \sum_{x=1}^{W}\sum_{y=1}^{H}S(x,y),
\end{equation}
where $W$ and $H$ are the width and height of the saliency map $S$, and $S(x,y)$ is the saliency score of the pixel at $(x,y)$.
%For binarization, we set all pixels with saliency value larger than $T_a$ as salient and the rest as unsalient.
We report the average precision, recall and F-measure over each dataset using this adaptive threshold.

Although commonly used, PR curves have limited value because they fail to consider true negative pixels. For a more balanced comparison, we adopt the mean absolute error (MAE) as another evaluation criterion. It is defined as the average pixelwise absolute difference between the binary ground truth ($G$) and the saliency map ($S$)~\cite{perazzi2012saliency},
\begin{equation}
MAE = \frac{1}{W\times H}\sum_{x=1}^{W}\sum_{y=1}^{H}|S(x,y) - G(x,y)|.
\end{equation}
MAE measures the numerical distance between the ground truth and the estimated saliency map, and is more meaningful in evaluating the applicability of a saliency model in a task such as object segmentation.

\subsection{Ablation Study} %Component-wise Efficacy
\subsubsection{Effectiveness of S-3CNN} As discussed in Section~\ref{sec:feature}, our multiscale CNN feature vector, S-3CNN, consists of three components, A, B and C. To show the effectiveness and necessity of these three parts, we have trained five additional models for comparison, which respectively take $feature$ A only, $feature$ B only, $feature$ C only, concatenated A and B, and concatenated A and C. These five models were trained on MSRA-B using the same setting as the one taking S-3CNN. Quantitative results were obtained on the testing images in the MSRA-B dataset. As shown in Fig.~\ref{fig:analysis_on_s3cnn}, the model trained using S-3CNN consistently achieves the highest PR curve and best performance on average precision, recall and F-measure. Models trained using two components perform much better than those trained using a single component.

These results demonstrate that the three components of our multiscale CNN feature vector are complementary to each other, and the training stage of our saliency model is capable of discovering and understanding region contrast information hidden in our multiscale features.

\subsubsection{Multilevel Decomposition} Our method exploits information from multiple levels of image segmentation. As shown in Fig. \ref{fig:analysis_on_multilayer}, the performance of a single segmentation level is not comparable to the performance of the fused model. The aggregated saliency map from 15 levels of image segmentation improves the average precision by $2.15\%$ and at the same time improves the recall rate by $3.47\%$ when it is compared with the result from the best-performing single level.

\subsubsection{Spatial Coherence}\label{sec:spatial_coherence}
In Section~\ref{sec:coherence}, a fully connected CRF model is incorporated to improve spatial coherence and refine the saliency scores obtained using MDF and HDHF. To validate its effectiveness, we have evaluated the performance of our saliency models with and without the CRF model across six datasets. As shown in Figure.~\ref{fig:spatial_coherence}, the CRF can consistently improve the results computed using MDF and HDHF across all the six datasets. Especially on the DUT-OMRON dataset which contains the largest number of testing images, the CRF increases the F-measure by 4.7\% on the HDHF results and 4.2\% on the MDF results.

\subsection{Evaluation on Contemporary CNN Architectures}
We evaluate the effectiveness of deep features extracted using different CNN architectures. We extract deep features using 4 contemporary deep CNN architectures, and train our saliency model on MSRA-B using such deep features. Evaluated CNN architectures include AlexNet~\cite{krizhevsky2012imagenet}, VGG16~\cite{simonyan2014very}, VGG19~\cite{simonyan2014very} and the R-CNN model~\cite{girshick2014rich}. We obtain quantitative comparison results on the testing images of the MSRA-B dataset. As shown in Table~\ref{tab:model_comparison}, the R-CNN model achieved slightly better performance than others. This model can better capture the feature of an image region probably because it was fine-tuned on a dataset of image regions for the purpose of object detection. 

We have also tried to jointly fine-tune the deep CNN model with our proposed MLP. As shown in Table~\ref{tab:model_comparison}, models with parameters fine-tuned can deliver slightly better results. Though our proposed model can effectively mine the contrast information from different scales of image regions and learn a superior deep contrast feature for saliency detection, it can hardly fine-tune much better description for each scale. In fact, all of these deep models are capable of capturing the feature of an image region but the regional feature performance of all these deep models does not vary much when applied in our contrast learning framework. To sum up, for saliency estimation, discovering the contrast between a region and its surrounding neighborhood is equally important. 

\begin{table}[]

\centering
\resizebox{0.45\textwidth}{!}
{
% \begin{tabular}{l|lll}
% \hline
% Deep Model & \multicolumn{1}{c}{AUC}       & \multicolumn{1}{c}{MAE}       & \multicolumn{1}{c}{F-measure} \\
% \hline
% RCNN       & {\color[HTML]{FE0000} \textbf{0.9778}} & {\color[HTML]{FE0000} \textbf{0.0658}} & {\color[HTML]{FE0000} \textbf{0.8884}} \\
% AlexNet    & 0.9745                        & 0.0701                        & 0.8785                        \\
% VGG16      & 0.9758                        & 0.0702                        & 0.8810                        \\
% VGG19      & 0.9769                        & 0.0692                        & 0.8818

% \end{tabular}

\begin{tabular}{l|lll}
Deep Model & AUC                                   & MAE                                   & F-Measure                             \\
\hline
RCNN       & {\color[HTML]{333333} 0.978}          & {\color[HTML]{333333} 0.066}          & {\color[HTML]{333333} 0.888}          \\
RCNN$^*$      & {\color[HTML]{FE0000} \textbf{0.979}} & {\color[HTML]{FE0000} \textbf{0.065}} & {\color[HTML]{FE0000} \textbf{0.901}} \\
\hline
AlexNet    & 0.975                                 & 0.070                                 & 0.879                                 \\
AlexNet$^*$   & 0.975                                 & 0.068                                 & 0.881                                 \\
\hline
VGG16      & 0.976                                 & 0.070                                 & 0.881                                 \\
VGG16$^*$     & 0.978                                 & 0.068                                 & 0.883                                 \\
\hline
VGG19      & 0.977                                 & 0.069                                 & 0.882                                 \\
VGG19$^*$     & 0.978                                 & 0.069                                 & 0.883                                
\end{tabular}

}
\caption{Comparison of saliency detection performance using different CNN architectures. $*$ refers to deep model with parameters fine-tuned.}
\label{tab:model_comparison}
\end{table}

\begin{figure}[ht]
    \centerline{
    \includegraphics[width = 0.248\textwidth]{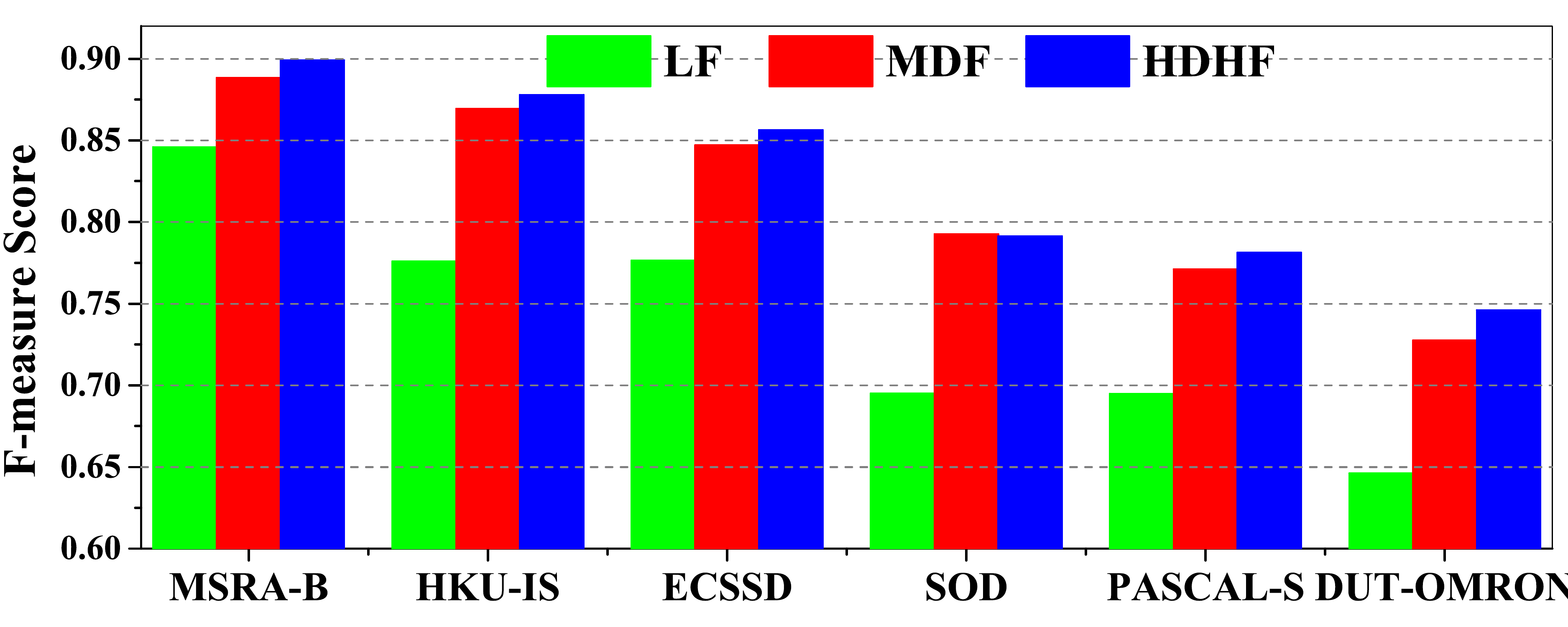}\hfill
    \includegraphics[width = 0.248\textwidth]{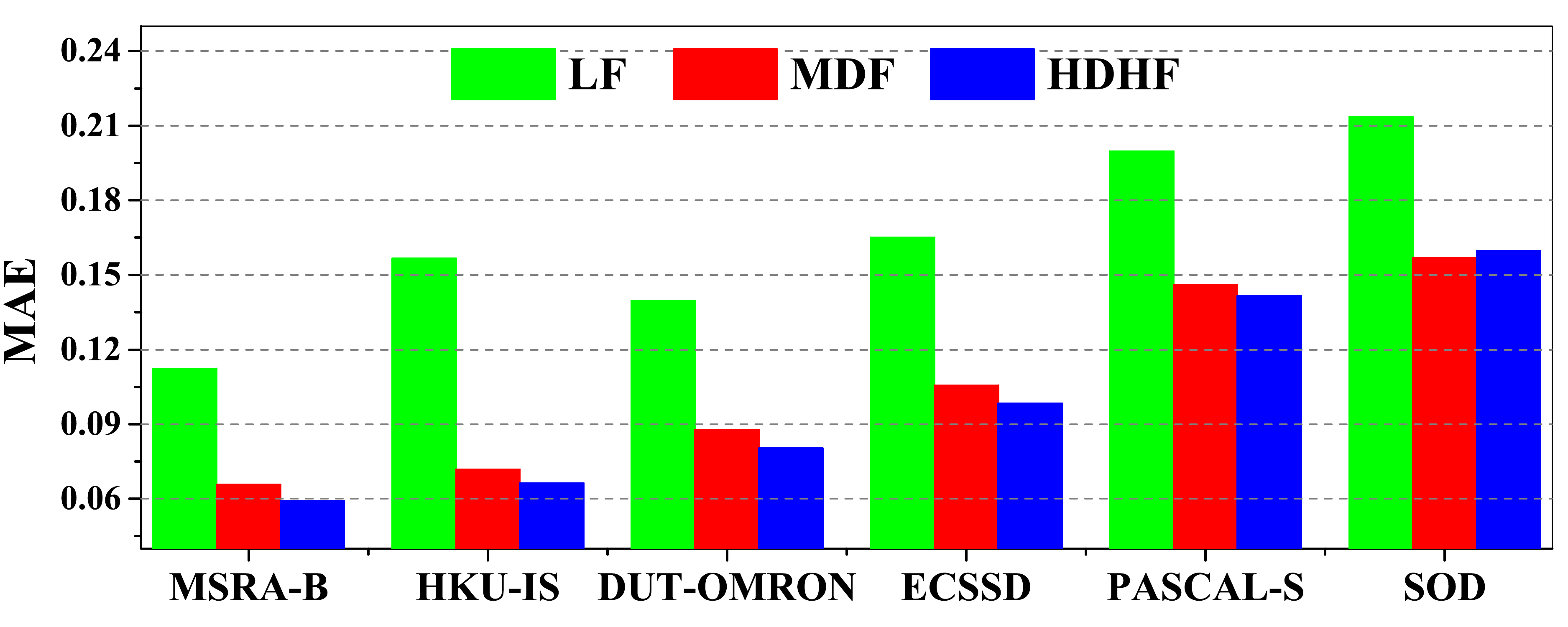}
  }\vspace{-0mm}
    \centerline{\hfill (a) \hfill\hfill (b) \hfill}\vspace{-1mm}
    \caption{Performance of our HDHF-based model. (a)Maximum F-measure of HDHF, MDF and LF on six saliency detection datasets. (b) MAE of HDHF, MDF and LF on the same datasets.}
\label{fig:mdf_vs_hdcf}
\end{figure}

\subsection{The Performance of HDHF}\label{sec:per_hdcf}
We evaluate the effectiveness of HDHF quantitatively by comparing its performance against that of MDF, which is based on deep features (S-3CNN) only, and LF, which is based on the 39-dimensional handcrafted low-level features only. Figure~\ref{fig:mdf_vs_hdcf} shows the F-measure and MAE of these three methods on six datasets. HDHF performs better than MDF most of the time, and consistently and significantly outperforms LF. Especially on the DUT-OMRON dataset, HDHF improves the F-measure of MDF by 2.6\% and LF by 12.4\% while, at the same time, lowers the MAE of MDF by 8.5\% and LF by 46.3\%.

\begin{figure*}[ht]
\begin{center}
   \includegraphics[width=0.95\textwidth]{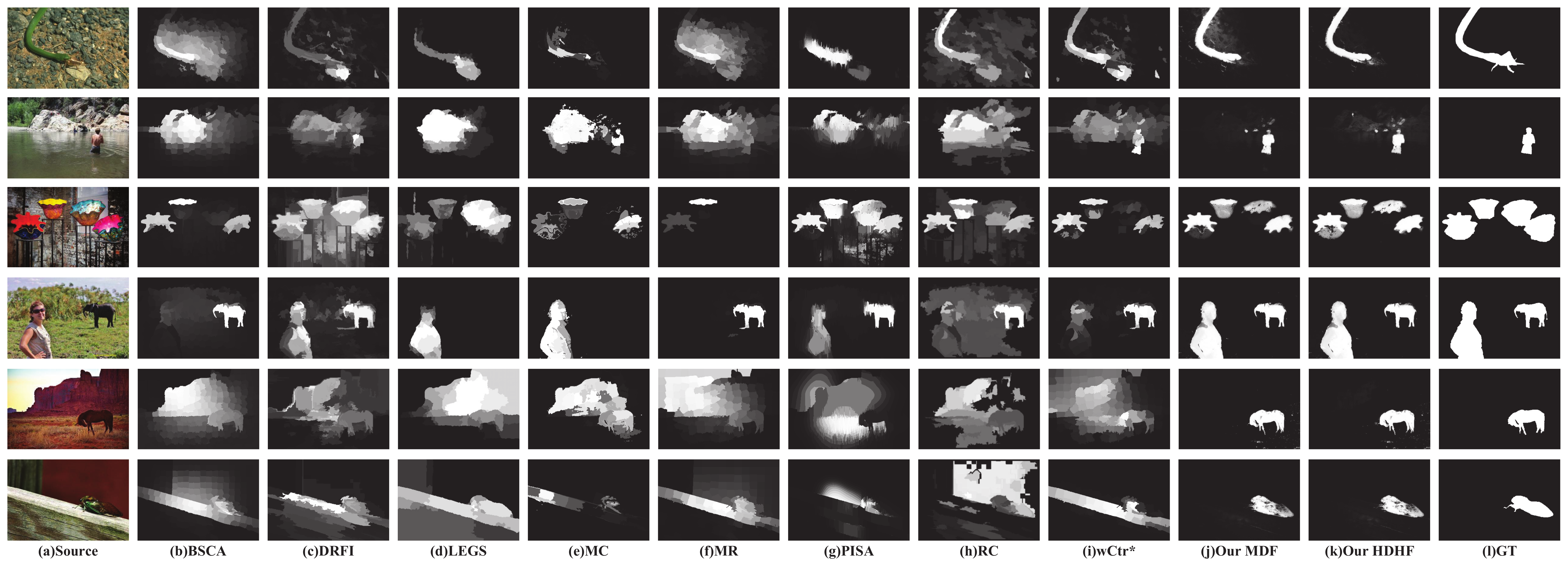}\vspace{-4mm}
\end{center}
   \caption{Visual comparison of saliency maps generated from 10 state-of-the-art methods, including our two models MDF and HDHF. The ground truth (GT) is shown in the last column. MDF and HDHF consistently produce saliency maps closest to the ground truth. %We compare MDF and HDHF against single-layer cellular automata~(BSCA)~\cite{qin2015saliency}, discriminative regional feature integration (DRFI)~\cite{jiang2013salient}, manifold ranking~(MR)~\cite{yang2013saliency}, pixelwise image saliency aggregating~(PISA)~\cite{wang2015pisa}, region based contrast~(RC)~\cite{ChengPAMI}, and optimized weighted contrast (wCtr$^*$)~\cite{zhu2014saliency}.
   }
\label{fig:long}
\end{figure*}

\begin{figure*}[t]
    \centerline{
    \includegraphics[width = 0.28\textwidth]{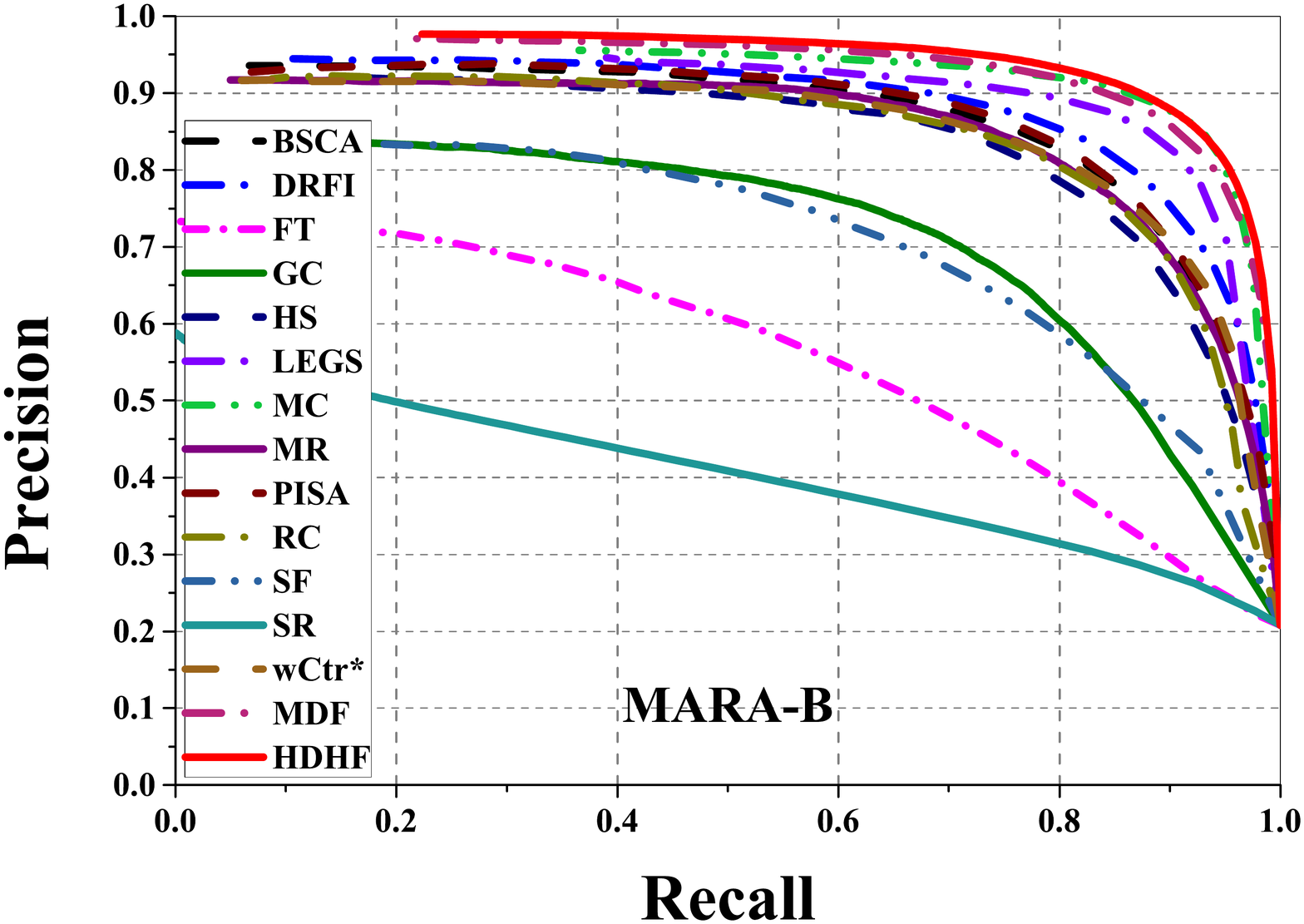}\hfill
    \includegraphics[width = 0.28\textwidth]{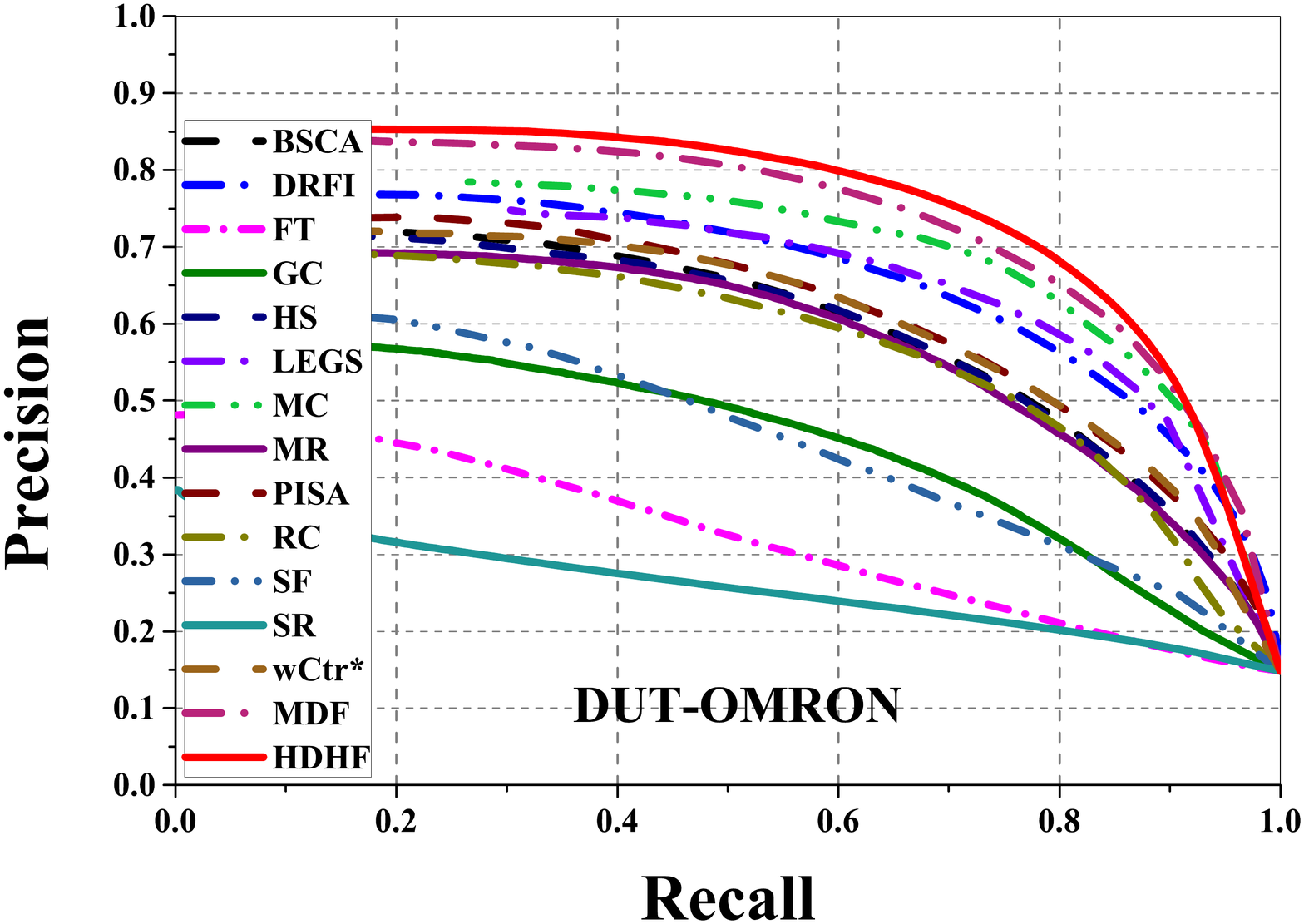}\hfill
    \includegraphics[width = 0.28\textwidth]{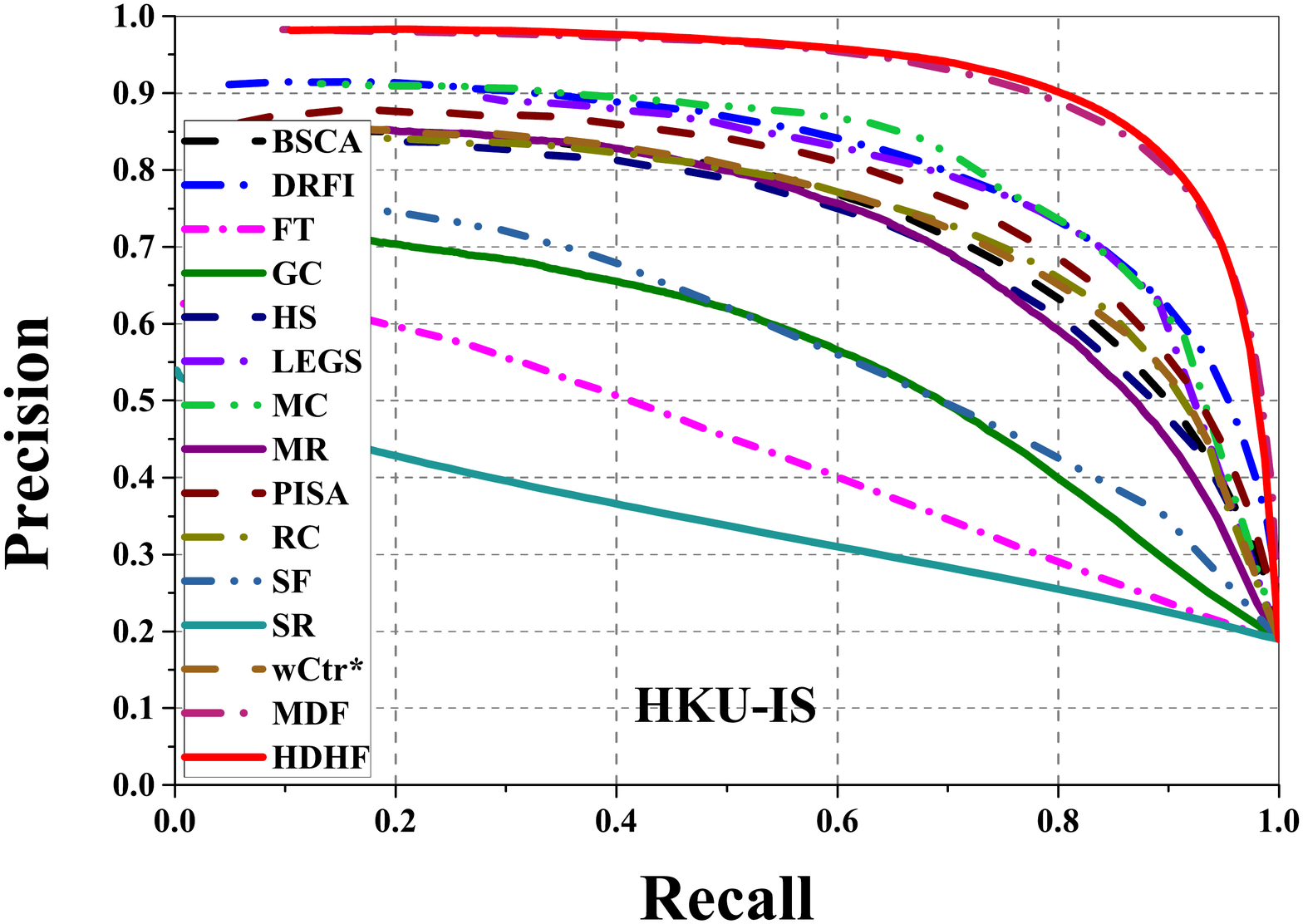}
  }
    %\centerline{
    %\includegraphics[width = 0.33\textwidth]{graphs/ecssd_pr_withdeep.eps}\hfill
    %\includegraphics[width = 0.33\textwidth]{graphs/pascal-s_pr_withdeep.eps}\hfill
    %\includegraphics[width = 0.33\textwidth]{graphs/sod_pr_withdeep.eps}
  %}
  \caption{Comparison of precision-recall curves of 15 saliency detection methods on 3 datasets. Our MDF and HDHF based models consistently outperform other methods across all the testing datasets. %Note that MC~\cite{zhao2015saliency} and LEGS~\cite{wang2015deep} are overrated on the MSRA-B dataset and LEGS~\cite{wang2015deep} is also overrated on the PASCAL-S dataset.
  }
  \label{fig:comps_pr}
\end{figure*}

% \begin{figure*}[t]
%     \centerline{
%     \includegraphics[width = 0.333\textwidth]{graphs/msra_roc_withdeep.eps}\hfill
%     \includegraphics[width = 0.333\textwidth]{graphs/dut-omron_roc_withdeep.eps}\hfill    
%     \includegraphics[width = 0.333\textwidth]{graphs/hku-is_roc_withdeep.eps}
%   }
%     \centerline{
%     \includegraphics[width = 0.333\textwidth]{graphs/ecssd_roc_withdeep.eps}\hfill
%     \includegraphics[width = 0.333\textwidth]{graphs/pascal-s_roc_withdeep.eps}\hfill
%     \includegraphics[width = 0.333\textwidth]{graphs/sod_roc_withdeep.eps}
%   }
%   \caption{Comparison of ROC curves of 15 saliency detection methods on 6 datasets. Our MDF and HDHF based models consistently outperform other methods across all the testing datasets. %Note that MC~\cite{zhao2015saliency} and LEGS~\cite{wang2015deep} are overrated on the MSRA-B dataset and LEGS~\cite{wang2015deep} is also overrateed on the PASCAL-S dataset.
%   }
%   \label{fig:comps_roc}
% \end{figure*}

\begin{figure*}[t]
    \centerline{
    \includegraphics[width = 0.28\textwidth]{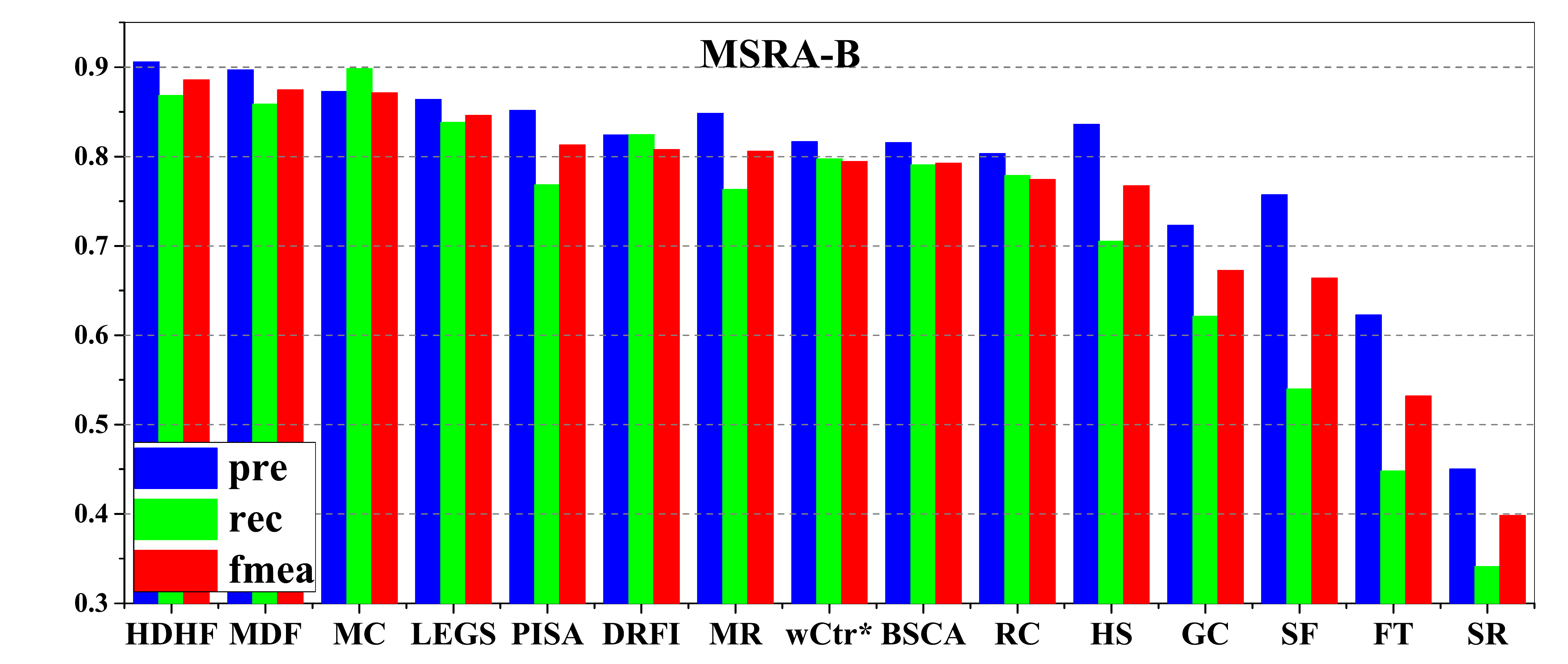}\hfill
    \includegraphics[width = 0.28\textwidth]{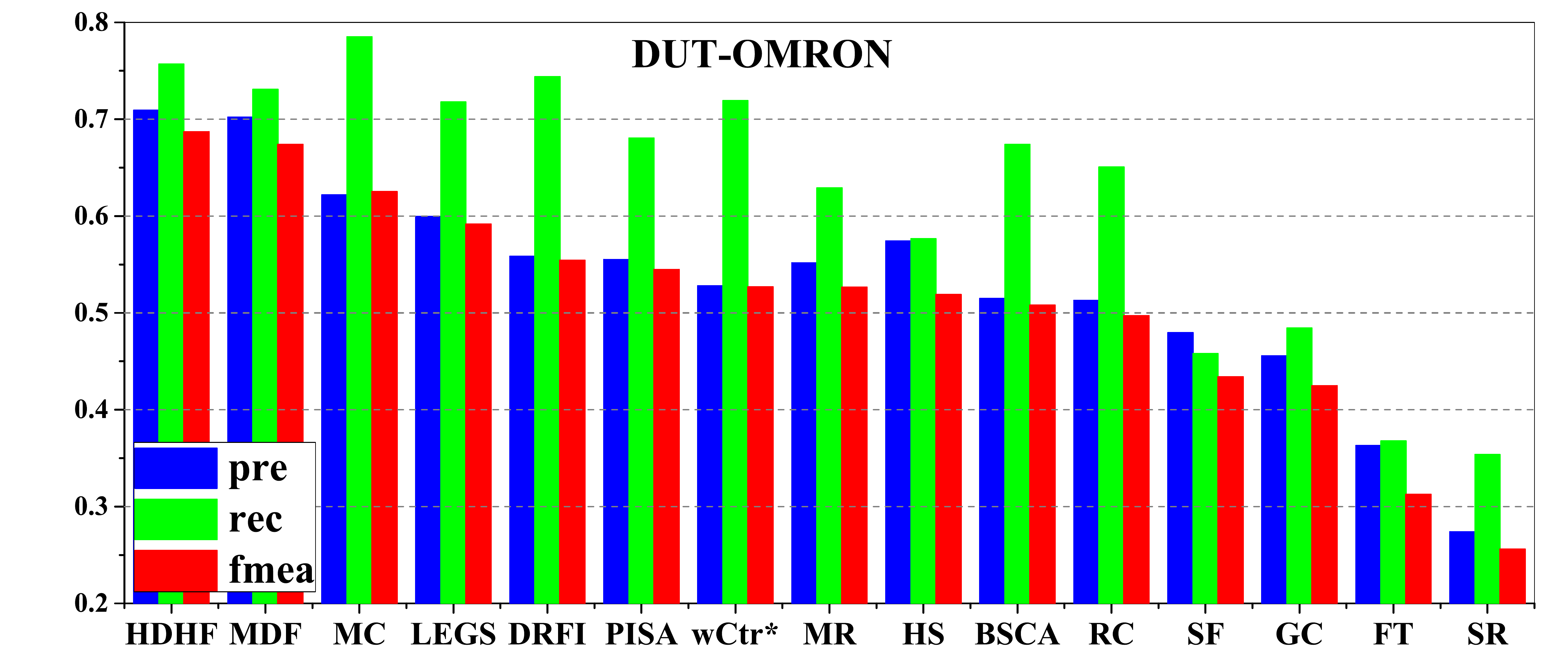}\hfill    
    \includegraphics[width = 0.28\textwidth]{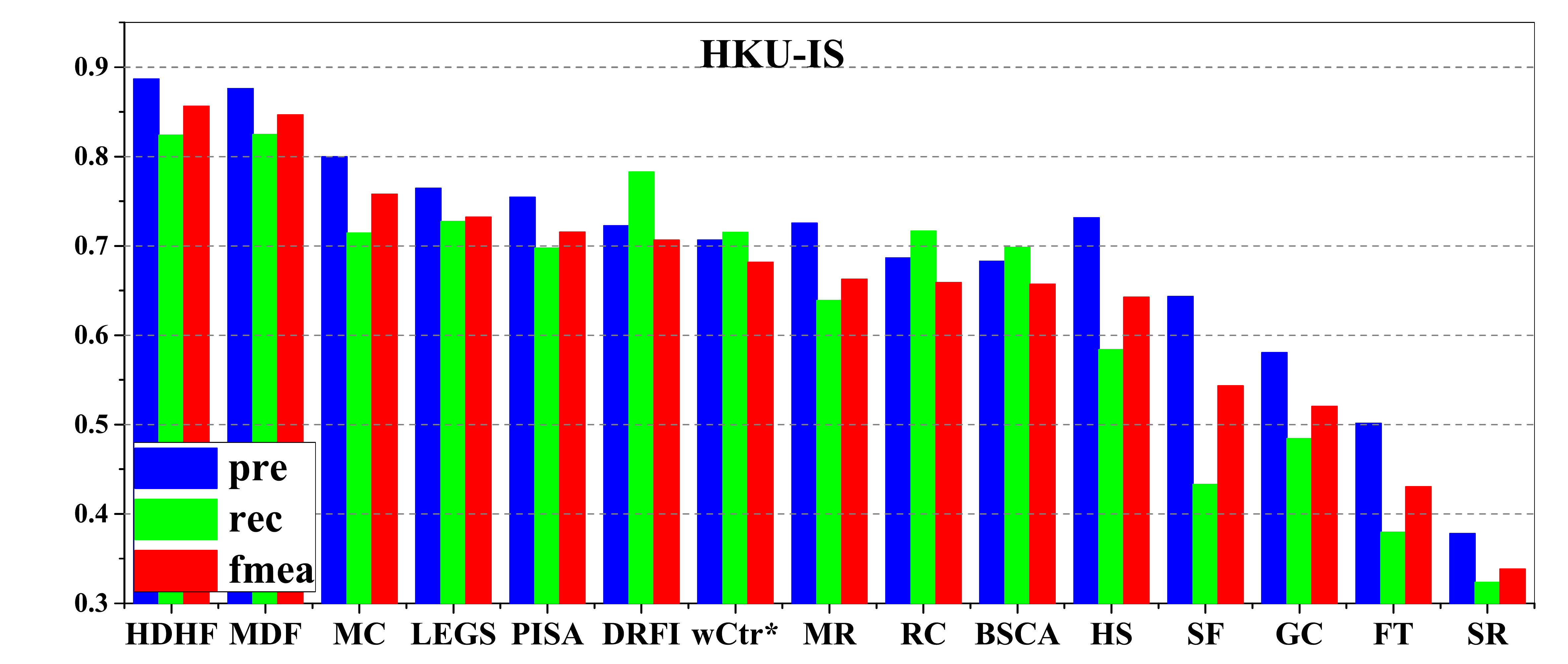}
  }
    %\centerline{
    %\includegraphics[width = 0.33\textwidth]{graphs/ecssd_prf_withdeep.eps}\hfill
    %\includegraphics[width = 0.33\textwidth]{graphs/pascal-s_prf_withdeep.eps}\hfill
    %\includegraphics[width = 0.33\textwidth]{graphs/sod_prf_withdeep.eps}
  %}
  \caption{Comparison of precision, recall and F-measure (computed using a per-image adaptive threshold) among 15 different methods on 3 datasets. %Our proposed MDF and HDCF consistently occupies the top two F-measure score across all the testing datasets.
  %Note that MC~\cite{zhao2015saliency} and LEGS~\cite{wang2015deep} are overrated on the MSRA-B dataset, and LEGS~\cite{wang2015deep} is also overrated on the PASCAL-S dataset.
  }\label{fig:comps_prf}
\end{figure*}

% \usepackage{multirow}

\subsection{Comparison with the State of the Art}
Let us compare our two saliency models (MDF and HDHF) with a number of existing state-of-the-art methods, including multi-context deep learning~(MC)~\cite{zhao2015saliency}, local estimation and global search based deep network~(LEGS)~\cite{wang2015deep}, single-layer cellular automata~(BSCA)~\cite{qin2015saliency}, pixelwise image saliency aggregating~(PISA)~\cite{wang2015pisa}, discriminative regional feature integration (DRFI)~\cite{jiang2013salient}, optimized weighted contrast (wCtr$^*$)~\cite{zhu2014saliency}, manifold ranking~(MR)~\cite{yang2013saliency}, global cues~(GC)~\cite{cheng2013efficient}, region based contrast~(RC)~\cite{ChengPAMI}, hierarchical saliency~(HS)~\cite{yan2013hierarchical}, saliency filters~(SF)~\cite{perazzi2012saliency}, frequency-tuned saliency~(FT)~\cite{achanta2009frequency} and the spectral residual approach (SR)~\cite{hou2007saliency}. For GC, RC, FT and SR, we use the implementation provided by the authors of \cite{ChengPAMI}; for other methods, we use their original implementation with recommended parameter settings.

A visual comparison is given in Fig.~\ref{fig:long}. For space consideration, we only choose the top 8 among all the methods we compare with for this visual demonstration. As can be seen, our models (Fig.\ref{fig:long}j\&k) perform well in a variety of challenging cases, e.g., cluttered background~(the first two rows%5-{th} and 6-{th} rows
), multiple disconnected salient objects~(3-{rd} and 4-{th} rows), low contrast between salient object and background (5-{th} and 6-{th} rows), and objects touching the image boundary (1-{st} and 4-{th} rows). In all the complex scenarios shown in Fig.~\ref{fig:long}, it is obvious that our models are able to successfully highlight entire salient objects, yielding saliency maps closest to the ground truth.

% A visual comparison is given in Fig.~\ref{fig:long}. For space consideration, we only choose the top 8 among all the methods we compare with for this visual demonstration. As can be seen, our models (Fig.\ref{fig:long}h\&i) perform well in a variety of challenging cases, e.g., cluttered background~(the first two rows%5-{th} and 6-{th} rows
% ), multiple disconnected salient objects~(3-{rd} and 4-{th} rows), low contrast between salient object and background (5-{th} and 6-{th} rows), and objects touching the image boundary (1-{st} and 4-{th} rows). In all the complex scenarios shown in Fig.~\ref{fig:long}, it is obvious that our models are able to successfully highlight entire salient objects, yielding saliency maps closest to the ground truth.

%A visual comparison is given in Fig.~\ref{fig:long}. For space consideration, we only choose the top 8 among all the methods we compare with for this visual demonstration. As can be seen, our models (Fig.\ref{fig:long}j\&k) perform well in a variety of challenging cases, e.g., low contrast between salient object and background (the first two rows), cluttered background (7-{th} and 8-{th} rows), multiple disconnected salient objects (11-{th} and 12-{th} rows) and objects touching the image boundary (7-{th} and 12-{th} rows). In all the complex scenarios shown in Fig.~\ref{fig:long}, it is obvious that our models are able to successfully highlight entire salient objects, yielding saliency maps closest to the ground truth.

As part of the quantitative evaluation, we first evaluate our method using Precision-Recall and ROC curves. As shown in Figs.~\ref{fig:comps_pr}% and \ref{fig:comps_roc}
, our methods (HDHF and MDF) consistently occupy the top two spots and outperform others on all benchmark datasets. %Refer to the supplemental document for ROC curves. 
The AUC~(Area under ROC curve) is reported in Table~\ref{tab:comp_quantity}. It is necessary to point out that the performance of MC~\cite{zhao2015saliency} is overrated on the MSRA-B dataset and the performance of LEGS~\cite{wang2015deep} is overrated on both the MSRA-B dataset and the PASCAL-S dataset because most images in the corresponding datasets were actually training samples for the publicly available trained models of MC~\cite{zhao2015saliency} and LEGS~\cite{wang2015deep} used in our comparison.

Precision, recall and F-measure values using the aforementioned adaptive threshold are shown in Fig.~\ref{fig:comps_prf}. Our method also achieves the highest precision and F-measure on all datasets. On the DUT-OMRON dataset, HDHF achieves 70.9\% in precision and 75.7\% in recall while the second best (MC) achieves 62.2\% in precision and 78.5\% in recall. Though the recall rate of MC is higher than ours, its precision is much lower. Thus it is much more likely for MC to misclassify unsalient pixels as salient ones. This is also reflected in the lower F-measure and higher MAE achieved with MC. Performance improvement becomes more obvious on HKU-IS. Compared with the second best (MC), our method increases the F-measure from 0.76 to 0.86, and achieves an increase of 10.9\% in precision while at the same time improving the recall by 15.3\%. %Similar conclusions can be made on most other datasets as well.

A quantitative comparison is shown in Table~\ref{tab:comp_quantity}. As can be seen, %it improves by 1.6\%~(ignore MC, DRFI is considered), 1.3\%, 4.7\%, 0.6\% and 3.8\% over the best-performing state-of-the-art algorithm according to the AUC scores on MSRA-B, ECSSD, HKU-IS, DUT-OMRON, PASCAL-S and SOD, respectively.
our HDHF based model improves the F-measure achieved by the best-performing existing algorithm by 6.4\%, 2.3\%, 10.0\%, 6.1\%, 5.5\% and 8.1\% respectively on MSRA-B (skipping MC and LEGS on this dataset), ECSSD, HKU-IS, DUT-OMRON, PASCAL-S (skipping LEGS on this dataset) and SOD. And at the same time, our HDHF based model also outperforms other existing methods in terms of MAE, which provides a better estimation of the visual difference between the predicted saliency map and the ground truth. As shown in Table~\ref{tab:comp_quantity}, our HDHF based model lowers the MAE by 48.0\%, 2.0\%, 35.3\%, 9.1\%, 2.1\% and 12.3\% respectively on MSRA-B (skipping MC and LEGS on this dataset), ECSSD, HKU-IS, DUT-OMRON, PASCAL-S (skipping LEGS on this dataset) and SOD.

In summary, the improvement our method achieves over the state of the art is substantial if we keep in mind the already good performance of state-of-the-art algorithms. Furthermore, the more challenging the dataset, the more obvious the advantages because our multiscale CNN features are capable of identifying subtle contrast among different parts of an image. More importantly, although our models are learned using the training set of the MSRA-B dataset, they are consistently among the top performers over all other challenging datasets.

\begin{table*}[t]
\centering
\resizebox{0.90\textwidth}{!}
{
\begin{tabular}{c||c|c|c|c|c|c|c|c|c|c|c|c|c|c|c|c}
\hline
Data Set                    & Metric    & BSCA  & DRFI  & FT    & GC    & HS    & LEGS                                  & MC                                    & MR    & PISA  & RC    & SF    & SR    & wCtr* & MDF                                   & HDHF                                  \\ \hline
                            & AUC       & 0.954 & 0.966 & 0.766 & 0.863 & 0.930 & 0.958                                 & {\color[HTML]{32CB00} \textbf{0.975}} & 0.941 & 0.954 & 0.937 & 0.886 & 0.710 & 0.948 & {\color[HTML]{3531FF} \textbf{0.978}} & {\color[HTML]{FE0000} \textbf{0.982}} \\ \cline{2-17}
                            & F-measure & 0.830 & 0.845 & 0.579 & 0.719 & 0.813 & 0.870                                 & {\color[HTML]{3531FF} \textbf{0.894}} & 0.824 & 0.837 & 0.817 & 0.700 & 0.430 & 0.820 & {\color[HTML]{32CB00} \textbf{0.888}} & {\color[HTML]{FE0000} \textbf{0.899}} \\ \cline{2-17}
\multirow{-3}{*}{MSRA-B}    & MAE       & 0.130 & 0.112 & 0.241 & 0.159 & 0.161 & 0.081                                 & {\color[HTML]{3531FF} \textbf{0.054}} & 0.127 & 0.102 & 0.138 & 0.166 & 0.224 & 0.110 & {\color[HTML]{32CB00} \textbf{0.066}} & {\color[HTML]{FE0000} \textbf{0.053}} \\ \hline
                            & AUC       & 0.922 & 0.943 & 0.663 & 0.767 & 0.885 & 0.925                                 & {\color[HTML]{32CB00} \textbf{0.948}} & 0.888 & 0.921 & 0.893 & 0.793 & 0.632 & 0.896 & {\color[HTML]{3531FF} \textbf{0.957}} & {\color[HTML]{FE0000} \textbf{0.960}} \\ \cline{2-17}
                            & F-measure & 0.758 & 0.782 & 0.430 & 0.597 & 0.727 & 0.827                                 & {\color[HTML]{32CB00} \textbf{0.837}} & 0.736 & 0.764 & 0.738 & 0.548 & 0.376 & 0.716 & {\color[HTML]{3531FF} \textbf{0.847}} & {\color[HTML]{FE0000} \textbf{0.856}} \\ \cline{2-17}
\multirow{-3}{*}{ECSSD}     & MAE       & 0.183 & 0.170 & 0.289 & 0.233 & 0.228 & 0.118                                 & {\color[HTML]{3531FF} \textbf{0.100}} & 0.189 & 0.150 & 0.186 & 0.219 & 0.264 & 0.171 & {\color[HTML]{32CB00} \textbf{0.106}} & {\color[HTML]{FE0000} \textbf{0.098}} \\ \hline
                            & AUC       & 0.911 & 0.950 & 0.710 & 0.777 & 0.884 & 0.907                                 & {\color[HTML]{32CB00} \textbf{0.928}} & 0.870 & 0.925 & 0.903 & 0.828 & 0.674 & 0.910 & {\color[HTML]{3531FF} \textbf{0.971}} & {\color[HTML]{FE0000} \textbf{0.972}} \\ \cline{2-17}
                            & F-measure & 0.723 & 0.776 & 0.477 & 0.588 & 0.710 & 0.770                                 & {\color[HTML]{32CB00} \textbf{0.798}} & 0.714 & 0.753 & 0.726 & 0.590 & 0.373 & 0.726 & {\color[HTML]{3531FF} \textbf{0.869}} & {\color[HTML]{FE0000} \textbf{0.878}} \\ \cline{2-17}
\multirow{-3}{*}{HKU-IS}    & MAE       & 0.174 & 0.167 & 0.244 & 0.211 & 0.213 & 0.118                                 & {\color[HTML]{32CB00} \textbf{0.102}} & 0.174 & 0.127 & 0.165 & 0.173 & 0.220 & 0.140 & {\color[HTML]{3531FF} \textbf{0.072}} & {\color[HTML]{FE0000} \textbf{0.066}} \\ \hline
                            & AUC       & 0.882 & 0.931 & 0.682 & 0.757 & 0.860 & 0.885                                 & {\color[HTML]{32CB00} \textbf{0.929}} & 0.853 & 0.893 & 0.859 & 0.810 & 0.688 & 0.894 & {\color[HTML]{FE0000} \textbf{0.935}} & {\color[HTML]{FE0000} \textbf{0.935}} \\ \cline{2-17}
                            & F-measure & 0.617 & 0.664 & 0.381 & 0.495 & 0.616 & 0.669                                 & {\color[HTML]{32CB00} \textbf{0.703}} & 0.610 & 0.630 & 0.599 & 0.495 & 0.298 & 0.630 & {\color[HTML]{3531FF} \textbf{0.728}} & {\color[HTML]{FE0000} \textbf{0.746}} \\ \cline{2-17}
\multirow{-3}{*}{DUT-OMRON} & MAE       & 0.191 & 0.150 & 0.250 & 0.218 & 0.227 & 0.133                                 & {\color[HTML]{3531FF} \textbf{0.088}} & 0.187 & 0.141 & 0.189 & 0.147 & 0.181 & 0.144 & {\color[HTML]{3531FF} \textbf{0.088}} & {\color[HTML]{FE0000} \textbf{0.080}} \\ \hline
                            & AUC       & 0.872 & 0.899 & 0.627 & 0.727 & 0.838 & 0.891                                 & {\color[HTML]{32CB00} \textbf{0.907}} & 0.852 & 0.866 & 0.840 & 0.746 & 0.671 & 0.866 & {\color[HTML]{3531FF} \textbf{0.921}} & {\color[HTML]{FE0000} \textbf{0.922}} \\ \cline{2-17}
                            & F-measure & 0.666 & 0.690 & 0.413 & 0.539 & 0.641 & {\color[HTML]{32CB00} \textbf{0.752}} & 0.740                                 & 0.661 & 0.660 & 0.644 & 0.493 & 0.392 & 0.655 & {\color[HTML]{3531FF} \textbf{0.771}} & {\color[HTML]{FE0000} \textbf{0.781}} \\ \cline{2-17}
\multirow{-3}{*}{PASCAL-S}  & MAE       & 0.224 & 0.210 & 0.309 & 0.266 & 0.264 & 0.157                                 & {\color[HTML]{3531FF} \textbf{0.145}} & 0.223 & 0.196 & 0.227 & 0.240 & 0.263 & 0.201 & {\color[HTML]{32CB00} \textbf{0.146}} & {\color[HTML]{FE0000} \textbf{0.142}} \\ \hline
                            & AUC       & 0.843 & 0.890 & 0.607 & 0.692 & 0.817 & 0.836                                 & {\color[HTML]{32CB00} \textbf{0.868}} & 0.812 & 0.848 & 0.828 & 0.714 & 0.679 & 0.827 & {\color[HTML]{3531FF} \textbf{0.899}} & {\color[HTML]{FE0000} \textbf{0.901}} \\ \cline{2-17}
                            & F-measure & 0.654 & 0.699 & 0.441 & 0.526 & 0.646 & {\color[HTML]{32CB00} \textbf{0.732}} & 0.727                                 & 0.636 & 0.660 & 0.657 & 0.516 & 0.444 & 0.653 & {\color[HTML]{FE0000} \textbf{0.793}} & {\color[HTML]{3531FF} \textbf{0.791}} \\ \cline{2-17}
\multirow{-3}{*}{SOD}       & MAE       & 0.251 & 0.223 & 0.323 & 0.284 & 0.283 & 0.195                                 & {\color[HTML]{32CB00} \textbf{0.179}} & 0.259 & 0.223 & 0.242 & 0.267 & 0.291 & 0.229 & {\color[HTML]{FE0000} \textbf{0.157}} & {\color[HTML]{3531FF} \textbf{0.160}} \\ \hline

\end{tabular}
}
\caption{Comparison of quantitative results including AUC (larger is better), maximum F-measure (larger is better) and MAE (smaller is better). The best three results are shown in \color[HTML]{FE0000}\textbf{red}\color{black}, \color[HTML]{3531FF}\textbf{blue}\color{black}, \color{black} and \color[HTML]{32CB00}\textbf{green} \color{black}color, respectively. Note that MC~\cite{zhao2015saliency} and LEGS~\cite{wang2015deep} are overrated on the MSRA-B dataset and LEGS~\cite{wang2015deep} is overrated on the PASCAL-S dataset.}
\label{tab:comp_quantity}
\end{table*}

\subsection{Efficiency}
While it takes around 20 hours to train our deep neural network based prediction model using the training set of the MSRA-B dataset, it only takes around 4 seconds to detect salient objects in a testing image with $400\times 300$ pixels on a PC with two NVIDIA GTX Titan Black GPUs and a 3.4GHz Intel processor using our MATLAB code. Noted that feature extraction efficiency can be improved using multi-GPU techniques provided in the latest Caffe framework~\cite{jia2014caffe}.

\section{Conclusions and Future Work}\label{sec:conclusion}

In this paper, we have introduced a neural network architecture, which has fully connected layers on top of CNNs responsible for feature extraction at three different scales. The proposed neural network architecture works as a feature learning model which deduces high-level semantic contrast and contextual relationships among the three scales. The penultimate layer of our neural network has been confirmed to be a very discriminative high-level feature vector for saliency detection and is complementary to handcrafted low-level features. To generate a more robust feature, we integrate low-level features with our deep contrast feature and feed the concatenated feature vector into a random forest regressor which maps the feature vector of each region to a saliency score. We aggregate multiple saliency maps computed for different levels of image segmentation to reduce error due to imperfect segmentation, and further incorporate a pixel-level CRF model to enhance spatial coherence. To promote further research and evaluation of visual saliency models, we have also constructed a large dataset of 4447 challenging images and their pixelwise saliency annotations. Experimental results demonstrate that our proposed method significantly outperforms all existing saliency estimation techniques on all public datasets.

%As future work, we are considering to train a saliency model in a fully convolutional mode.

As future work, we are considering to improve the efficiency of deep feature extraction. In this paper, we treat each region as an independent unit in feature extraction without any shared computation. 
%Though parallel computing is applied in our implementation, the current time complexity is still far from real-time performance. 
We are considering spatial pyramid pooling networks (SPPnets)~\cite{he2014spatial} for speeding up regional feature extraction by computing a single convolutional feature map for an entire image and then extracting all regional features from this shared feature map. We are also considering the application of our deep contrast feature in other pixel labeling problems, e.g. depth prediction from monocular images, eye-fixations and object proposals.

% or


%\appendices

%\section{}

\ifCLASSOPTIONcompsoc
  \section*{Acknowledgments}

\else
  \section*{Acknowledgment}
\fi

The authors would like to thank Sai Bi, Wei Zhang, and Feida Zhu for their help during the construction of our dataset. The first author is supported by Hong Kong Postgraduate Fellowship.

\ifCLASSOPTIONcaptionsoff
  \newpage
\fi



%\bibliographystyle{IEEEtran}

{\small

\bibliography{ref,saliency}{}
\bibliographystyle{IEEEtran}
}


\begin{IEEEbiography}[{\includegraphics[width=1in,height=1.25in,clip,keepaspectratio]{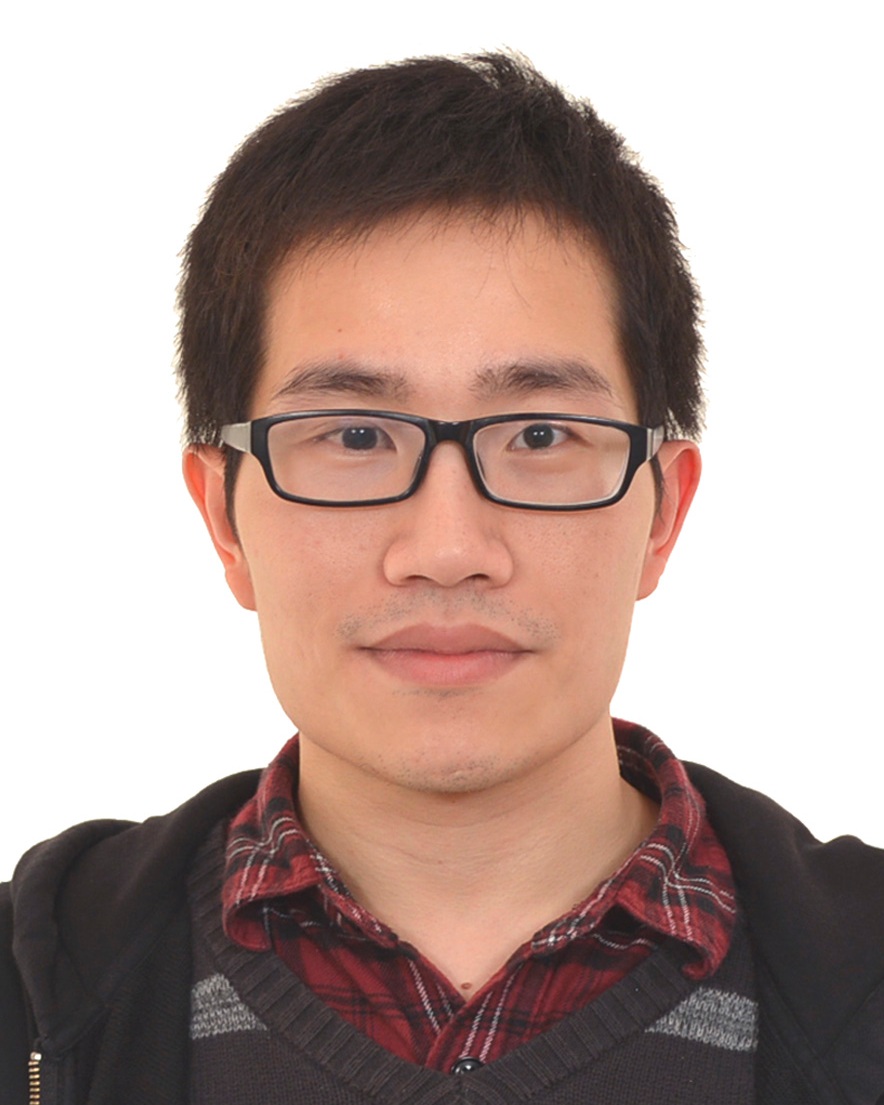}}]{Guanbin Li}
received the PhD degree from the University of Hong Kong in 2016. He is currently a research scientist at Sun-Yat Sen University. %BE and Master’s degree in computer science from Sun-Yat Sen University in 2009 and 2012.  He is currently a PhD candidate in the Department of Computer Science, the University of Hong Kong. 
He is a recipient of Hong Kong Postgraduate Fellowship. His current research interests include computer vision, image processing, and deep machine learning.
\end{IEEEbiography}

\begin{IEEEbiography}[{\includegraphics[width=1in,height=1.25in,clip,keepaspectratio]{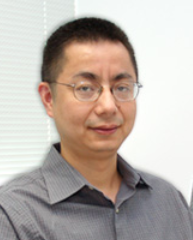}}]{Yizhou Yu}
received the PhD degree from University of California at Berkeley in 2000. He is currently a professor at The University of Hong Kong, and was a faculty member at University of Illinois at Urbana-Champaign for twelve years. He received 2002 US National Science Foundation CAREER Award, and 2007 NNSF China Overseas Distinguished Investigator Award. Prof Yu has served on the editorial board of IEEE Transactions on Visualization and Computer Graphics, The Visual Computer, and International Journal of Software and Informatics. He has also served on the program committee of many leading international conferences, including SIGGRAPH, SIGGRAPH Asia, and International Conference on Computer Vision. His current research interests include deep learning methods for visual computing, digital geometry processing, video analytics and biomedical data analysis.
\end{IEEEbiography}

%\end{IEEEbiography}


%\newpage



%\vfill


\end{document}